\documentclass{article} 
\usepackage{iclr2025_conference,times}

\usepackage{amsmath,amsfonts,bm}

\def\eqref#1{equation~\ref{#1}}

\def\1{\bm{1}}

\DeclareMathAlphabet{\mathsfit}{\encodingdefault}{\sfdefault}{m}{sl}
\SetMathAlphabet{\mathsfit}{bold}{\encodingdefault}{\sfdefault}{bx}{n}

\usepackage[utf8]{inputenc} 
\usepackage[T1]{fontenc}    
\usepackage{hyperref}       
\usepackage{url}            
\usepackage{booktabs}       
\usepackage{amsfonts}       
\usepackage{nicefrac}       
\usepackage{microtype}      
\usepackage{xcolor}         
\usepackage{graphicx}
\usepackage{hyperref}
\usepackage{amsmath}
\usepackage{cleveref}
\usepackage{color, colortbl}
\usepackage{tabularray}
\usepackage{multirow}
\usepackage{nicematrix}
\usepackage{caption}
\usepackage{subcaption}
\usepackage{booktabs} 
\usepackage{wrapfig}

\definecolor{mygray}{gray}{0.9}
\definecolor{cvprblue}{rgb}{0.21,0.49,0.74}
\hypersetup{pdfborder={0 0 0}}
\hypersetup{colorlinks=true}
\hypersetup{citecolor=cvprblue}

\title{VILA-U: a Unified Foundation Model Integrating Visual Understanding and Generation}

\newcommand{\abb}{VILA-U }
\newcommand{\abbns}{VILA-U}

\definecolor{revision_table}{rgb}{0,0.7,1}
\definecolor{revision}{rgb}{0,0.3,1}
\definecolor{mydarkgreen}{rgb}{0.02,0.6,0.02}
\definecolor{darkred}{rgb}{0.459,0.0,0.08}
\definecolor{darkorange}{rgb}{0.40,0.2,0.02}
\definecolor{darkpurple}{RGB}{111,0,255}
\definecolor{myred}{rgb}{1.0,0.0,0.0}
\definecolor{mygold}{rgb}{0.75,0.6,0.12}
\definecolor{mydarkgray}{rgb}{0.66, 0.66, 0.66}

\author{%
  Yecheng Wu$^{1,2}$\thanks{ Equal Contribution.} \quad
  Zhuoyang Zhang$^{2}$\footnotemark[1] \ \thanks{Part of the work done during an internship at NVIDIA.} \quad
  Junyu Chen$^{1,2}$ \quad
  Haotian Tang$^{2}$\footnotemark[2] \quad \\
  \textbf{Dacheng Li}$^{4}$\footnotemark[2] \quad
  \textbf{Yunhao Fang}$^{5}$\footnotemark[2] \quad
  \textbf{Ligeng Zhu}$^{3}$ \quad
  \textbf{Enze Xie}$^{3}$ \quad \\
  \textbf{Hongxu Yin}$^{3}$ \quad
  \textbf{Li Yi}$^{1}$ \quad
  \textbf{Song Han}$^{2,3}$ \quad
  \textbf{Yao Lu}$^{3}$ \\
  Tsinghua University$^{1}$\quad MIT$^{2}$\quad NVIDIA$^{3}$\quad UC Berkeley$^{4}$\quad UC San Diego$^{5}$\\
  \textcolor{darkred}{\url{https://hanlab.mit.edu/projects/vila-u}}
}

\iclrfinalcopy 
\begin{document}

\maketitle

\begin{abstract}
\textbf{VILA-U} is a \textbf{U}nified foundation model that integrates \textbf{V}ideo, \textbf{I}mage, \textbf{La}nguage understanding and generation. Traditional visual language models (VLMs) use separate modules for understanding and generating visual content, which can lead to misalignment and increased complexity. In contrast, VILA-U employs a single autoregressive next-token prediction framework for both tasks, eliminating the need for additional components like diffusion models. This approach not only simplifies the model but also achieves near state-of-the-art performance in visual language understanding and generation. The success of VILA-U is attributed to two main factors: the unified vision tower that aligns discrete visual tokens with textual inputs during pretraining, which enhances visual perception, and autoregressive image generation can achieve similar quality as diffusion models with high-quality dataset. This allows VILA-U to perform comparably to more complex models using a fully token-based autoregressive framework. Our code is open sourced at \textcolor{darkred}{\url{https://github.com/mit-han-lab/vila-u}}.
\end{abstract}

\section{Introduction}
\label{sec:introduction}

In recent years, large language models (LLMs) have demonstrated superior capabilities in various language tasks. Their appealing properties like instruction following, zero-shot generalization, and few-shot in-context learning motivate researchers to combine them with vision models to build visual language models (VLMs) for multi-modal tasks. Many efforts \citep{instructblip, llava, lin2023vila} in this field have achieved remarkable performance on visual language understanding. In these works, visual inputs are projected onto LLMs' semantic space through a vision model like CLIP \citep{clip} to bridge two modalities by including text-image alignment objectives.

In addition to visual understanding, another essential research direction in combining visual and language modalities is visual generation. There are two popular approaches for text-guided image generation. One approach employs diffusion models~\citep{ldm}, a powerful tool for various generation tasks. The other line of work converts visual content into discrete tokens through vector quantization (VQ) and then leveraging autoregressive transformers for high-quality and diverse generation~\citep{VQGAN,vit-vqgan,lee2022autoregressive, var, llamagen}. 


Witnessing the rapid advancements in both visual understanding and generation, an emerging trend is to unify these techniques into a single multi-modal framework. 
Prior to VILA-U, there are two main approaches to achieving such unification: (1) One approach \citep{lwm, CM3Leon, xie2024show} utilizes a VQGAN-based \citep{VQGAN} tokenizer to convert visual inputs into discrete tokens and leverages an autoregressive model for both understanding and generation. However, \citep{xie2024show} has shown that visual tokens from VQGAN-based encoder lack semantic information and usually results in a severe performance drop in downstream visual understanding tasks. (2) Another approach \citep{zhan2024anygpt, seed-llama, jin2023unified} utilizes a codebook to quantize features produced by a pre-trained vision model like CLIP. Since CLIP features encode rich semantic information, these approaches generally achieve significantly better performance on understanding tasks. However, these tokenizers lack decoding capability, requiring an external visual generation model, such as a diffusion model, to use the generated visual tokens as conditions for producing visual outputs. This approach adds complexity to infrastructure design. Available large-scale foundation model training pipelines and deployment systems have already been highly optimized for language modeling with next-token prediction. 
Designing and maintaining an additional stack to support diffusion models would incur significant engineering costs. 

In this work, we present \textbf{\abbns}, an \textit{end-to-end autoregressive} framework with a unified next-token prediction objective for both visual and text inputs that can achieve competitive performance on both visual language understanding and generation tasks, without the help of external components like diffusion models.
We identify two critical principles to unify vision and language modalities: (1) Existing unified end-to-end autoregressive VLMs cannot achieve competitive visual understanding performance because the discrete VQGAN tokens are trained solely on image reconstruction loss and are not aligned with textual inputs. Therefore, it is crucial to introduce text alignment during VQ vision tower pretraining to enhance perception capabilities. (2) Autoregressive image generation can attain similar quality as diffusion models if trained on high-quality data with sufficient size. Guided by these insights, VILA-U features a unified foundation vision tower that converts visual inputs into discrete tokens through vector quantization and aligns these tokens with textual inputs using contrastive learning. The multi-modal training of VILA-U takes advantage of a unified next-token prediction objective for both visual and textual tokens on a small-size high-quality image-text corpus.

We evaluate \abb on common visual language tasks, including image-language understanding, video-language understanding, image generation and video generation. VILA-U significantly narrows the gap in visual understanding performance between end-to-end autoregressive models and continuous-token VLMs, while introducing competitive \textit{native} visual generation capabilities.

\section{Related Work}
\label{sec:related_work}

\looseness=-1
\textbf{Large Language Models (LLMs).} LLMs based on pre-trained large-scale transformers \citep{NIPS2017Transformer} has drastically revolutionized natural language processing field. Featuring gigantic model size and pre-training data corpus, LLM has achieved remarkable performance on various
linguistic tasks. The development of open-source LLMs such as LLaMA \citep{llama}, Mixtral \citep{mixtral} and Vicuna \citep{vicuna} has furthered nourished research on how to adopt LLM for complex language tasks. Besides excellent zero-shot generalizability to diverse domains, LLM is commonly finetuned on custom datasets for better performance on specific tasks. Instruction tuning \citep{ChatGPT, chung2024scaling, ouyang2022training} also stands as a key step for better outputs in applying LLMs. In this work, we adopt the LLaMA-2-7B \citep{llama} model as our basic LLM.

\looseness=-1
\textbf{Visual Language Models (VLMs).} Combining computer vision and natural language processing gives rise to VLM in this LLM era. In VLMs, researchers leverage vision foundation models such as CLIP \citep{clip}, BLIP \citep{blip} and CoCa \citep{CoCa} to extract visual features, align with texts, and feed them into LLM to achieve the cross-modality understanding between texts and visual content. Building upon such progress, many VLMs \citep{alayrac2022flamingo, blip2, llava, lin2023vila, luo2024deem, tian2024mm} have been designed and trained on extensive vision-language data to achieve remarkable performance on visual understanding and reasoning tasks. 
In this work, we aim to develop a VLM with visual understanding capacities comparable to prior works, while also possessing the new capacity of visual generation.

\textbf{Unified Visual Language Models.} Numerous efforts have been made to develop unified visual language models capable of generating both text and visual content, including images and videos. There are two mainstream methods to generate visual content in VLMs. 
Many works \citep{Emu, Emu2, jin2023unified, seed-llama, li2024mgm, seed-x, video-lavit, ge2023planting} combine VLMs with diffusion models like Stable Diffusion \citep{ldm} for high-quality image generation. Other works \citep{lwm, CM3Leon, Unified-io2, chameleon, xie2024show} adopt VQGAN-based vision encoders to convert visual inputs into discrete tokens and make LLMs learn to predict them. For more details on the distinction between our method and other unified visual language models, please refer to Appendix~\ref{appendix:A}.


\section{Methods}
\label{sec:methods}

\begin{figure}[t]
    \centering
    \includegraphics[width=.95\textwidth]{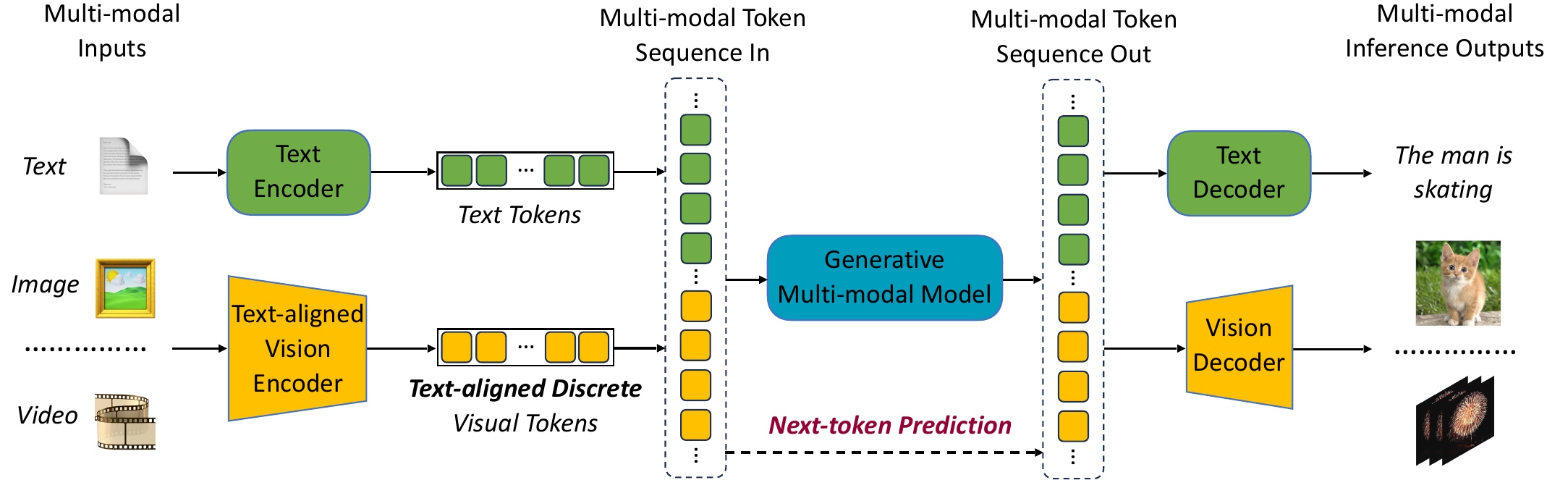}
    \caption{\textbf{An overview of our framework's multi-modal training and inference process.} Visual inputs are tokenized into discrete tokens and concatenated with textual tokens to form a multi-modal token sequence. All tokens are involved in our next-token prediction process, enabling a unified training objective. During inference, the output tokens are decoded by our text detokenizer or vision tower decoder to yield multi-modal content. }
    \label{fig:unified_model}
    \vspace{-15pt}
\end{figure}

This work proposes a multi-modal framework that aims to unify visual and language modalities effectively. The key components enabling such unification are a unified foundation vision tower that converts visual inputs into discrete tokens aligned with text, and a unified multi-modal generative training procedure. An overview of the main multi-modal training and inference process within our framework is depicted in Figure \ref{fig:unified_model}.

\subsection{Unified Foundation Vision Tower}
\label{subsec:vision_tower}

\begin{figure}[t]
    \centering
    \includegraphics[width=0.93\textwidth]{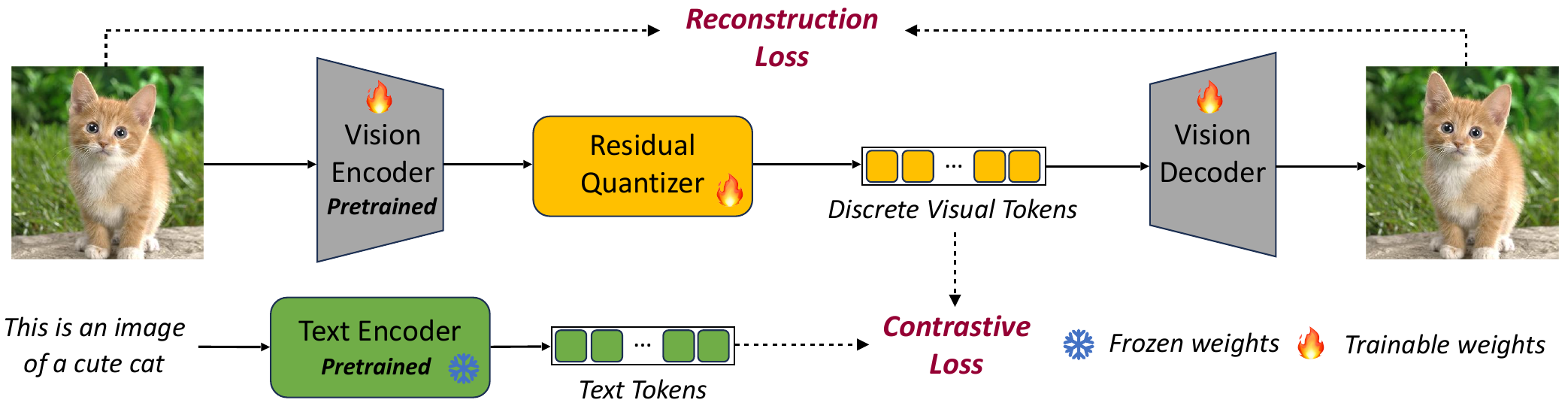}
    \caption{
    \looseness=-1
    \textbf{Overview of our unified foundation vision tower.} Given input images the features extracted by the vision encoder are discretized using residual quantization. Then the discrete vision features are meanwhile put into the vision decoder to reconstruct images and used to perform the text-image alignment. During this process, the reconstruction loss and contrastive loss are computed to update the vision tower, endowing it to produce discrete visual features with text alignment.}
    \label{fig:vision_tower}
    \vspace{-15pt}
\end{figure}

To support diverse visual understanding and generation tasks, we first build a unified foundation vision tower to provide appropriate visual features. We propose to include text-image contrastive loss and VQ-based image reconstruction loss in our vision tower training, empowering the text alignment and discrete tokenization abilities for our vision tower. As depicted in Figure~\ref{fig:vision_tower}, the features extracted from images are primarily discretized through residual quantization. Then in one route, the discrete visual features are fed into a decoder to reconstruct the image and compute the reconstruction loss; on the other route, we compute the image-text contrastive loss between the discrete visual features and the textual features provided by a text encoder. With this training procedure, the vision tower learns to extract discrete features suitable for both understanding and generation in our VLM.

\vspace{-6pt}
\paragraph{Unified Training Recipe.} 
Training the unified vision tower with two objectives from scratch would be difficult, because alignment and reconstruction tasks require high-level semantic and low-level appearance features, respectively. Training the entire vision tower from scratch with both objectives could induce conflicting goals. In practice, we observe that training the vector-quantized vision tower from scratch with both image reconstruction and contrastive loss results in a mere 5\% Top-1 accuracy for zero-shot image classification on ImageNet~\citep{deng2009imagenet} after several epochs of training. 

To address this issue, we experiment with different training recipes (failed recipes are listed in Appendix~\ref{appendix:C}) and find the following solution to be most effective. Instead of learning both objectives simultaneously, we suggest first equipping the model with text-image alignment ability and then learning reconstruction while maintaining alignment ability. We initialize the vision encoder and text encoder with pretrained weights from the CLIP model to ensure good text-image alignment. Next, we freeze the text encoder and keep all vision components trainable using both contrastive and reconstruction loss. The contrastive loss maintains alignment ability, while the reconstruction loss develops reconstruction ability. This approach converges quickly and yields strong performance. The pre-trained CLIP weights contain learned high-level priors, which are difficult and computationally expensive to learn from scratch. Initializing with these weights enables the binding of low-level and high-level features much faster and more tractably for the vision encoder. With this recipe, we can train a vision tower that exhibits both good text alignment and image reconstruction abilities. We use weighted sum to combine the text-image contrastive loss and VQ-based image reconstruction loss:

\vspace{-0.5cm}
\begin{equation}
\begin{aligned}
    \mathcal{L}_{total} = w_{contra}\mathcal{L}_{contra} +  w_{recon}\mathcal{L}_{recon}\\
\end{aligned}
\end{equation}

In our experiments, we pick $w_{contra}$ = 1 and $w_{recon}$ = 1.

\paragraph{Residual Vector Quantization.} Our visual features are discretely quantized, so their representation ability heavily depends on the code size used in our quantizer. Since we hope they contain both high-level and low-level features, we need more capacities in their vector feature space, making a larger code size necessary for good performance in downstream tasks. However, too many codes for each image will result in too many tokens for LLM to produce in the visual generation process, incurring much latency.
So in an attempt to increase the vector feature capacity and meanwhile maintain a reasonable number of tokens for LLM, we adopt a residual vector quantization method following RQ-VAE \citep{lee2022autoregressive} to discretize a vector $\mathbf{z}$ as $D$ discrete codes:
\begin{equation}
\begin{aligned}
    \mathcal{R} \mathcal{Q}(\mathbf{z} ; \mathcal{C}, D)&=\left(k_1, \cdots, k_D\right) \in[K]^D, \\
\end{aligned}
\end{equation}
where $\mathcal{C}$ is the codebook, $K = |\mathcal{C}|$ and $k_d$ is the code of $\mathbf{z}$ at depth $d$. Starting with $\mathbf{r}_0 = \mathbf{z}$, we recursively perform vector quantization by
\begin{equation}
\begin{aligned}
    k_d &= \mathcal{Q}\left(\mathbf{r}_{d - 1}, \mathcal{C}\right), \\
    \mathbf{r}_d &= \mathbf{r}_{d - 1} - \mathbf{e}\left(k_d\right),\\
\end{aligned}
\end{equation}
for each depth $d = 1, 2,\cdots, D$, where $\mathbf{e}$ is the codebook embedding table and $\mathcal{Q}$ is the standard vector quantization:
\begin{equation}
\begin{aligned}
    \mathcal{Q}(\mathbf{z} ; \mathcal{C})=\underset{k \in[K]}{\arg \min }\|\mathbf{z}-\mathbf{e}(k)\|_2^2.
\end{aligned}
\end{equation}
The quantized vector for $\mathbf{z}$ is the sum over the depth dim: $\widehat{\mathbf{z}} = \sum_{i = 1}^D\mathbf{e}\left(k_i\right)$. Intuitively, in each depth we choose a code to reduce the quantization error. So compared to the standard vector quantization methods, we have $D$ codes to quantize one vector, allowing for finer approximation and larger feature space. During multi-modal training and inference, LLM only needs to predict the code embedding, with codes in different depth sequentially produced by a depth transformer taking the code embedding as the initial input, as we will introduce in Section \ref{subsec:unified_model}. So with this residual quantization, we can enhance the representation capability of our vision tower while incurring little latency.  

\subsection{Unified Multi-modal Generative Pre-training}
\label{subsec:unified_model}

\looseness=-1
Figure \ref{fig:unified_model} presents an overview of our unified multi-modal pre-training process. Our vision tower encoder processes visual inputs sequentially, generating a 1D token sequence. This sequence is then concatenated with text tokens to form a multi-modal sequence. To distinguish between modalities and enable visual content generation, we insert special tokens: $\texttt{<image\_start>}$ and $\texttt{<image\_end>}$ at the start and end of image tokens, and $\texttt{<video\_start>}$ and $\texttt{<video\_end>}$ at the start and end of video tokens. Video tokens are the direct concatenation of multi-frame image tokens.

\textbf{Pre-training data form.} In terms of unified pre-training data, we leverage different concatenation forms between text and visual tokens to facilitate both understanding and generation. We use \texttt{[image, text]}, \texttt{[text, image]}, and \texttt{[text, video]} forms, with supervision loss added only on the latter modality in each pair to avoid unconditional content generation and promote modality alignment. We also employ an interleaved text and image concatenation form for enhanced understanding, with supervision loss applied solely to the text. Notably, we exclude the \texttt{[video, text]} form during pre-training for efficiency reasons, as we find incorporating it during supervised fine-tuning effectively yields excellent video understanding ability.

\textbf{Training Objective.} Since both visual tokens and text tokens are discrete, we can train our LLM with the general language modeling next-token prediction objective. However, due to the use of residual quantization for visual tokens, the training objectives for text and visual tokens differ slightly. For text tokens, the negative log-likelihood loss is calculated as

\begin{equation}
\begin{aligned}
    \mathcal{L}_\text{text} = -\sum_{i = 1}^T\log P_\theta\left(y_i|y_{<i}\right),
\end{aligned}
\end{equation}

where $T$ is the length of the multi-modal sequence and $i$ only counts when the text token appears at position $i$. 
For visual tokens, residual quantization introduces a depth-stacked structure of codes at each visual position $j$. To address this, we leverage the depth transformer introduced in RQ-VAE \citep{lee2022autoregressive}. Specifically, given the code embedding $h_j$ generated by the LLM for visual tokens at position $j$, the depth transformer autoregressively predicts D residual tokens  ($k_{j1}$, ..., $k_{jD}$). During training, the input of the depth transformer $v_{jd}$ at depth d is defined as the sum of the code embeddings of up to depth $d-1$ for $d > 1$ such that

\begin{equation}
\begin{aligned}
v_{jd} = \sum_{d'=1}^{d-1} \mathbf{e}(k_{jd'}),
\end{aligned}
\end{equation}
and $v_{j1} = h_j$. Thus, the depth transformer predicts the next code for a finer estimation of the feature $\boldsymbol{\hat{z}}_j$ based on the previous estimations up to $d-1$. Then the negative log-likelihood loss for visual tokens is

\vspace{-6pt}
\begin{equation}
\begin{aligned}
    \mathcal{L}_\text{visual} = -\sum_{j = 1}^T\sum_{d = 1}^D\log P_\delta\left(k_{jd}|k_{j,<d}\right),
\end{aligned}
\end{equation}

where  $T$ is the length of the multi-modal sequence and $j$ only counts when a visual token appears at position $j$. During the multi-modal pre-training, the weights of the depth transformer are randomly initialized and updated together with the LLM.

\section{Experiments}
\label{sec:experiments}

In this section, we introduce comprehensive experiments to evaluate our method on various visual understanding and generation tasks. Firstly, we outline our experimental setup, including the model architecture, training datasets, and evaluation benchmarks. Subsequently, we evaluate the performance of our unified foundation vision tower. Then, we compare our method with other popular VLMs on various visual understanding and generation benchmarks. Finally, we give some qualitative results.

\subsection{Experimental Setup}
\label{subsec:setup}
In our experiments, we employ LLaMA-2-7B \citep{touvron2023llama} as our base language model. For the vision tower, we choose SigLIP-Large-patch16-256 / SigLIP-SO400M-patch14-384 \citep{zhai2023sigmoid} as our vision encoder architecture, and adopt the residual quantizer, depth transformer as well as the decoder architecture from RQ-VAE \citep{lee2022autoregressive}. The quantizer codebook size is 16384. All images and videos are resized to a resolution of $256\times 256$ / $384\times 384$, with each image or video frame converted into a $16\times 16\times 4$ / $27\times 27\times 16$ code with the residual depth $D = 4$ / $D = 16$. We train our vision tower on COYO-700M \citep{coyo-700m} and evaluate it for zero-shot classification and reconstruction performance on ImageNet \citep{imagenet}. For visual understanding, we leverage 1M \texttt{[image, text]} data from ShareGPT4V \citep{sharegpt4v}, 6M interleaved text and image data from MMC4 \citep{mmc4}. For visual generation, we incorporate 15M high-quality \texttt{[text, image]} data curated from our internal dataset and 1M \texttt{[text, video]} data from OpenVid \citep{nan2024openvid} datasets. Classifier-free guidance \citep{cfg} is employed for visual generation with a CFG value of 3.

For examining visual understanding ability, we evaluate our model on the widely adopted zero-shot image-based visual-language benchmarks including VQAv2 \citep{vqa_v2}, GQA \citep{gqa}, TextVQA \citep{textvqa}, POPE \citep{POPE}, MME \citep{mme}, SEED \citep{seed}, MM-Vet \citep{mm-vet} and video-based visual-language benchmarks including ActivityNet \citep{activitynet}, MSVD \citep{chen2011collecting}, MSRVTT \citep{MSVD}, TGIF \citep{tgif}.

To evaluate the visual generation capability, we use MJHQ-30K \citep{playgroundv2.5} and GenAI-Bench \citep{GenAI-Bench} for image generation and VBench \citep{vbench} for video generation. MJHQ-30K adopts the FID between generated images and 30K high-quality images to reflect the overall capability of image generation. GenAI-Bench is a challenging image-to-text generation benchmark that reflects the comprehensive generative abilities of image generation models. Vbench is a comprehensive benchmark suite for video generative models that decomposes the generation quality into multiple well-defined dimensions to facilitate fine-grained and objective evaluation.


\subsection{Unified Foundation Vision Tower}

We present the commonly used metrics reconstruction FID (rFID) and Top-1 accuracy for zero-shot image classification on ImageNet to measure the reconstruction and text alignment capabilities of the unified foundation vision tower in Table \ref{tab:vison_tower}. Please refer to the Appendix~\ref{appendix:1} for the qualitative reconstruction results. Our model achieves significantly better reconstruction results than VQ-GAN. Our rFID is slightly inferior to that of RQ-VAE when using the same code shape. This is expected as the introduction of contrastive loss during training, aimed at enhancing image understanding, led to a decrease in reconstruction quality. For the text alignment capability, our unified vision tower achieves a Top-1 accuracy of 73.3 / 78.0 under 256 / 384 resolution. This demonstrates the exceptional text alignment capability of our unified vision tower. However, it is worth noting that both the rFID and Top-1 accuracy of the vision tower only serves as a medium indicator. As the unified vision tower is an integral component of the entire autoregressive model, we believe that its performance on downstream tasks, such as visual understanding and generation, holds greater significance.


\begin{table}[ht]
\setlength{\tabcolsep}{5.5pt}
\centering
\caption{The reconstruction FID (rFID) and Top-1 accuracy for zero-shot image classification of our unified vision tower on ImageNet.}
\scalebox{0.8}{
\begin{tabular}{lccccc}
\toprule
\textbf{Model} & \textbf{Pretrained Weights} & \textbf{Resolution} & \textbf{Shape of Code} & \textbf{rFID}$\downarrow$ & \textbf{Top-1 Accuracy}$\uparrow$ \\
    \midrule
    VQ-GAN & -- & 256 $\times$ 256 & 16 $\times$ 16 & 4.98 & -- \\
    RQ-VAE & -- & 256 $\times$ 256 & 8 $\times$ 8 $\times$ 4 & 3.20 & -- \\
    RQ-VAE & -- & 256 $\times$ 256 & 16 $\times$ 16 $\times$ 4 & 1.30 & -- \\ 
    \midrule
    \rowcolor{mygray}Ours & SigLIP-Large  & 256 $\times$ 256 & 16 $\times$ 16 $\times$ 4 & 1.80 & 73.3 \\
    \rowcolor{mygray}Ours & SigLIP-SO400M & 384 $\times$ 384 & 27 $\times$ 27 $\times$ 16 & 1.25 & 78.0 \\
    \bottomrule
\end{tabular}
}
\label{tab:vison_tower}
\end{table}

\subsection{Quantitative Evaluation}

\begin{table}[t!]
\setlength{\tabcolsep}{5.5pt}
\centering
\caption{Comparison with leading methods on image-based visual language benchmarks. Our performance is close to leading VLMs, surpassing many methods by a large margin under the same LLM size, even with a discrete visual token type. * indicates that images in the training split of these datasets are observed during VLM training.}
\scalebox{0.72}{
\begin{tabular}{l l l l | c c c c c c c}
\toprule
{\bf Method} & {\bf LLM} & {\bf Visual Token} & {\bf Res.} & {\bf VQAv2} & {\bf GQA} & {\bf TextVQA} & {\bf POPE} & {\bf MME} & {\bf SEED} & {\bf MM-Vet}   \\
\midrule
LLaVA-1.5 & Vicuna-1.5-7B & Continuous & 336 & 78.5$^*$ & 62.0$^*$ & 58.2 & 85.9 & 1510.7 & 58.6 & 30.5 \\
VILA & LLaMA-2-7B & Continuous & 336 & 79.9$^*$ & 62.3$^*$ & 64.4 & 85.5 & 1533.0 & 61.1 & 34.9 \\
Unified-IO 2 & 6.8B from scratch & Continuous & 384 & 79.4$^*$ & -- & -- & 87.7 & -- & 61.8 & -- \\
InstructBLIP & Vicuna-7B & Continuous & 224 & -- & 49.2 & 50.1 & -- & -- & 53.4 & 26.2 \\
IDEFICS-9B & LLaMA-7B & Continuous & 224 & 50.9 & 38.4 & 25.9 & -- & -- & -- & -- \\
Emu & LLaMA-13B & Continuous & 224 & 52.0 & -- & -- & -- & -- & -- & -- \\
LaVIT & LLaMA-7B & Continuous & 224 & 66.0 & 46.8 & -- & -- & -- & -- & -- \\
DreamLLM & Vicuna-7B & Continuous & 224 & 72.9$^*$ & -- & 41.8 & -- & -- & -- & 36.6 \\
Video-LaVIT & LLaMA-2-7B & Continuous & 224 & 80.2$^*$ & 63.6$^*$ & -- & -- & 1581.5 & 64.4 & 35.0 \\
Emu2-Chat & Emu2-37B & Continuous & 448 & 84.9$^*$ & 65.1$^*$ & 66.6$^*$ & -- & -- & -- & -- \\
MM-Interleaved & Vicuna-13B & Continuous & 224 & 80.2$^*$ & 60.5$^*$ & 61.0 & -- & -- & -- & -- \\
DEEM & Vicuna-7B & Continuous & 448 & 68.2$^*$ & 55.7$^*$ & -- & -- & -- & -- & 37.4 \\ 
CM3Leon-7B & 7B from scratch & Discrete & 256 & 47.6 & -- & -- & -- & -- & -- & -- \\
LWM & LLaMA-2-7B & Discrete & 256 & 55.8 & 44.8 & 18.8 & 75.2 & -- & -- & 9.6 \\
Show-o & Phi-1.5-1.3B & Discrete & 256 & 59.3$^*$ & 48.7$^*$ & -- & 73.8 & 948.4 & -- & -- \\
SEED-LLaMA & Vicuna-7B & Discrete & 224 & 66.2 & -- & -- & -- & -- & 51.5 & -- \\
\midrule
\rowcolor{mygray}Ours & LLaMA-2-7B & Discrete & 256 & 75.3$^*$ & 58.3$^*$ & 48.3 & 83.9 & 1336.2 & 56.3 & 27.7 \\
\rowcolor{mygray}Ours & LLaMA-2-7B & Discrete & 384 & 79.4$^*$ & 60.8$^*$ & 60.8 & 85.8 & 1401.8 & 59.0 & 33.5 \\
\bottomrule
\end{tabular}
\vspace{-0.5cm}
}
\label{tab:image_language_understanding}
\end{table}
\begin{table}[t!]
\setlength{\tabcolsep}{5.5pt}
\centering
\caption{Comparison with leading methods on video-based visual language benchmarks. The performance of our method is close to state-of-the-art VLMs, surpassing many methods under the same LLM size, even with a discrete visual token type.}
\scalebox{0.72}{
\begin{tabular}{l l l l | c c c c}
\toprule
{\bf Method} & {\bf LLM} & {\bf Visual Token} & {\bf Res.}  & {\bf MSVD-QA} & {\bf MSRVTT-QA} & {\bf TGIF-QA} & {\bf Activity Net-QA}    \\
\midrule
Unified-IO 2 & 6.8B from scratch & Continuous & 384  & 52.1 & 42.5 & -- & --\\
Emu & LLaMA-13B & Continuous & 224 & -- & 18.8 & 8.3 & -- \\
VideoChat & Vicuna-7B & Continuous & 224  & 56.3 & 45 & 34.4 & -- \\
Video-LLaMA & LLaMA-2-7B & Continuous & 224  & 51.6 & 29.6 & -- & --\\
Video-ChatGPT & LLaMA-2-7B & Continuous & 224  & 64.9 & 49.3 & 51.4 & 35.2 \\
Video-LLava & Vicuna-7B & Continuous & 224 & 70.7 & 59.2 & 70.0 & 45.3 \\
Video-LaVIT & LLaMA-2-7B & Continuous & 224 & 73.5 & 59.5 & -- & 50.2 \\
Emu2-Chat & Emu2-37B & Continuous & 448 & 49.0 & 31.4 & -- & -- \\
LWM & LLaMA-2-7B & Discrete & 256 & 55.9 & 44.1 & 40.9 & -- \\SEED-LLaMA & Vicuna-7B & Discrete & 224 & 40.9 & 30.8 & -- & -- \\
\midrule
\rowcolor{mygray}Ours & LLaMA-2-7B & Discrete & 256  & 73.4 & 58.9 & 51.3 & 51.6\\
\rowcolor{mygray}Ours & LLaMA-2-7B & Discrete & 384  & 75.3 & 60.0 & 51.9 & 52.7\\
\bottomrule
\end{tabular}
}
\label{tab:video_language_understanding}
\end{table}

\textbf{Visual Understanding Tasks.} Table \ref{tab:image_language_understanding} and Table \ref{tab:video_language_understanding} summarize the comparison between our method and other leading VLMs on the image-language and video-language benchmarks respectively. Compared to the mainstream choice of continuous visual tokens produced by foundation models like CLIP, the VQGAN-based discrete visual tokens have less alignment with text, thus harming VLMs' performance on visual understanding tasks. With our unified foundation vision tower, our model can have a performance close to leading VLMs even with discrete visual tokens.

\begin{wraptable}{r}{7.5cm}
\centering
\resizebox{0.95\linewidth}{!}{
\begin{tabular}{lccc}
\toprule
\textbf{Method} & \textbf{Type} & \textbf{\#Images} & \textbf{FID}$\downarrow$ \\
    \midrule
    SD v2.1 & Diffusion & -- & 26.96 \\
    SD-XL & Diffusion & 2000M & 9.55 \\
    PixArt & Diffusion & 25M & 6.14 \\ 
    Playground v2.5 & Diffusion & -- & 4.48 \\ 
    LWM & Autoregressive & -- & 17.77 \\
    Show-o & Autoregressive & 36M & 15.18 \\
    \midrule
    \rowcolor{mygray}Ours (256) & Autoregressive & 15M & 12.81 \\
    \rowcolor{mygray}Ours (384) & Autoregressive & 15M & 7.69 \\
    \bottomrule
\end{tabular}
}
\caption{\small Comparison with other visual generation methods on MJHQ-30K evaluation benchmark.}
\label{tab:image_generation_fid}
\vspace{-0.4in}
\end{wraptable}

\begin{table}[t!]
\setlength{\tabcolsep}{5.5pt}
\centering
\caption{Comparison with other visual generation methods on GenAI-Bench \citep{GenAI-Bench}. The results show that our method outperforms previous autoregressive visual generation methods. For \textit{advanced} prompts that require better text following ability to generate, our method can have a relatively small performance gap with diffusion-based methods, even with much less training data.}
\scalebox{0.8}{
    \begin{NiceTabular}{lccccccc|c}
    \CodeBefore
    \Body
    \toprule[1.2pt]
    \multirow{2}{*}{\textbf{Method}} & \multirow{2}{*}{\textbf{Type}} & \multirow{2}{*}{\#\textbf{Training Images}} & \multirow{2}{*}{\bf Attribute{$\uparrow$}} & \multirow{2}{*}{\bf Scene{$\uparrow$}} & \multicolumn{3}{c}{\bf Relation{$\uparrow$}} & \multirow{2}{*}{\bf Overall$\uparrow$}   \\
    \cmidrule{6-8}
    &  &  & & & Spatial & Action & Part   \\
    \midrule
    SD v2.1 & Diffusion & 2000M & 0.80 & 0.79 & 0.76 & 0.77 & 0.80 & 0.78 \\
    SD-XL & Diffusion & 2000M & 0.84 & 0.84 & 0.82 & 0.83 & 0.89 & 0.83\\
    Midjourney v6 & Diffusion & -- & 0.88 & 0.87 & 0.87 & 0.87 & 0.91 & 0.87 \\
    DALL-E 3 & Diffusion & -- & 0.91 & 0.90 & 0.92 & 0.89 & 0.91 & 0.90 \\
    LWM & Autoregressive & -- & 0.63 & 0.62 & 0.65 & 0.63 & 0.70 & 0.63 \\
    Show-o & Autoregressive  & 36M & 0.72 & 0.72 & 0.70 & 0.70 & 0.75 & 0.70  \\
    \midrule
    \rowcolor{mygray}Ours (256) & Autoregressive & 15M & 0.78 & 0.78 & 0.77 & 0.78 & 0.79 & 0.76 \\
    \rowcolor{mygray}Ours (384) & Autoregressive & 15M & 0.75 & 0.76 & 0.75  & 0.73  & 0.75 & 0.73 \\
    \bottomrule[1.2pt]
    \end{NiceTabular}}\\
    \vspace{1mm}
    (a) VQAScores on \textit{basic} prompts of GenAI-Bench \\
    \vspace{1mm}
\scalebox{0.76}{
    \begin{NiceTabular}{lccccccc|c}
    \CodeBefore
    \Body
    \toprule[1.2pt]
    \multirow{2}{*}{\textbf{Method}} & \multirow{2}{*}{\textbf{Type}} & 
    \multirow{2}{*}{\#\textbf{Training Images}} & 
    \multirow{2}{*}{\bf Count{$\uparrow$}} & \multirow{2}{*}{\bf Differ{$\uparrow$}} & \multirow{2}{*}{\bf Compare{$\uparrow$}} & \multicolumn{2}{c}{\bf Logical{$\uparrow$}} & \multirow{2}{*}{\bf Overall{$\uparrow$}}   \\
    \cmidrule{7-8}
    &  &  & & &  & Negate & Universal   \\
    \midrule
    SD v2.1 & Diffusion & 2000M & 0.68 & 0.70 & 0.68 & 0.54 & 0.64 & 0.62 \\
    SD-XL & Diffusion & 2000M & 0.71 & 0.73 & 0.69 & 0.50 & 0.66 & 0.63 \\
    Midjourney v6 & Diffusion & -- & 0.78 & 0.78 & 0.79 & 0.50 & 0.76 & 0.69 \\
    DALL-E 3 & Diffusion & --
 & 0.82 & 0.78 & 0.82 & 0.48 & 0.80 & 0.70 \\
    LWM & Autoregressive & --
 &0.59 & 0.58 & 0.54 & 0.49 & 0.52 & 0.53 \\
   Show-o & Autoregressive & 36M & 0.70  & 0.62 & 0.71 & 0.51 & 0.65 & 0.60 \\
    \midrule
    \rowcolor{mygray}Ours (256) & Autoregressive & 15M & 0.70 & 0.71 & 0.74 & 0.53 & 0.66 & 0.64 \\
    \rowcolor{mygray}Ours (384) & Autoregressive & 15M & 0.68 & 0.67 & 0.71  & 0.51  & 0.64 & 0.61 \\
    \bottomrule[1.2pt]
    \end{NiceTabular}} \\
    \vspace{1mm}
     (b) VQAScores on \textit{advanced} prompts of GenAI-Bench
    \vspace{-10pt}
\label{tab:image_generation}
\end{table}

\looseness=-1
\textbf{Visual Generation Tasks.} As shown in Table \ref{tab:image_generation_fid}, \abb can achieve a better FID than other autoregressive methods and have comparable performance with some diffusion based methods. This result shows the feasibility of our method for visual generation. Table \ref{tab:image_generation} summarizes the quantitative results of our method and other visual generation methods on GenAI-Bench. Although Our method is inferior to diffusion-based visual generation methods that have been trained on billions-level image-text pairs, our method has comparable performance with SD v2.1 \citep{rombach2022high} and SD-XL \citep{sdxl} on \textit{advanced} prompts even trained with magnitude-level less data. This further shows that \abb can learn the correlation among visual and textual modalities effectively with our unified training framework. For video generation, we evaluate our method on VBench~\citep{vbench} and compare it against Open-Sora~\citep{zheng1open}, CogVideo~\citep{hong2022cogvideo}, and CogVideoX~\citep{yang2024cogvideox}. The results, presented in Table \ref{tab:vbench}, demonstrate that our method achieves performance that is better than CogVideo and comparable to Open-Sora, highlighting the effectiveness of our approach.

\begin{wraptable}{r}{8.5cm}
\centering
\resizebox{0.95\linewidth}{!}{
\begin{tabular}{lccc}
\toprule
\textbf{Method} & \textbf{Total Score}$\uparrow$ & \textbf{Quality Score}$\uparrow$ & \textbf{Semantic Score}$\uparrow$ \\
    \midrule
    Open-Sora & 75.91 & 78.82 & 64.28 \\
    CogVideo & 67.01 & 72.06 & 46.83 \\
    CogVideoX & 81.61 & 82.75 & 77.04 \\
    \midrule
    \rowcolor{mygray}Ours (256)& 74.01 & 76.26 & 65.04 \\
    \bottomrule
\end{tabular}
}
\caption{\small Comparison with other visual generation methods on VBench \citep{vbench}.}
\label{tab:vbench}
\vspace{-0.1in}
\end{wraptable}

\subsection{Qualitative Evaluation}

\begin{figure}[b]
    \centering
    \includegraphics[width=\textwidth]{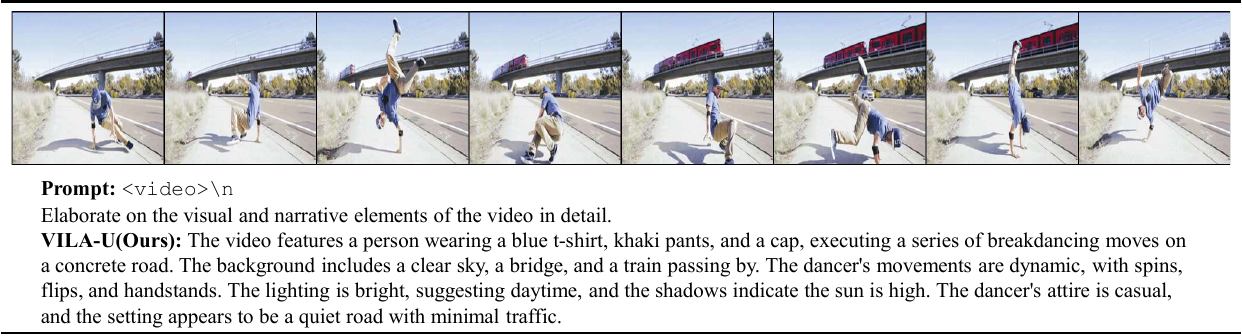}
    \caption{VILA-U can correctly caption videos and cover all the details, thanks to the text alignment of our vision encoder.}
    \label{fig:video_caption}
\end{figure}

\begin{figure}[htbp]
    \centering

    \begin{minipage}{0.47\textwidth}
        \centering
        \includegraphics[height=3.7cm]{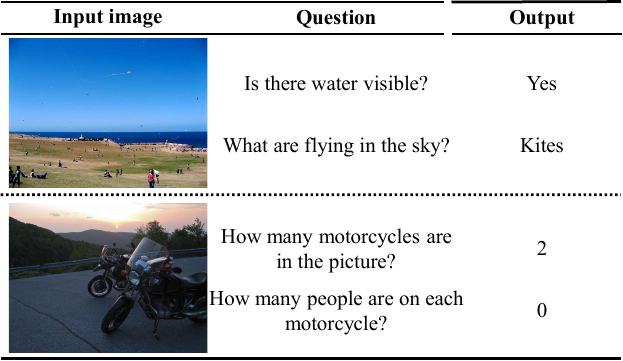}
        \caption{VILA-U has good visual question answering capability. The images and questions are from the test split of VQAv2 dataset.}
        \label{fig:vqa}
    \end{minipage}\hfill
    \begin{minipage}{0.51\textwidth}
        \centering
        \includegraphics[height=3.7cm]{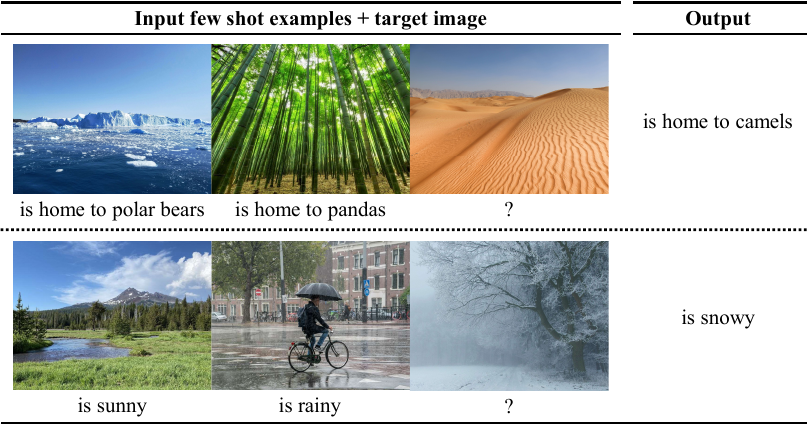}
        \caption{VILA-U has good in-context learning capability. We feed two image-text pairs and a third image as the context to prompt the VLM.}
        \label{fig:icl}
    \end{minipage}
    
\end{figure}

\looseness=-1
\textbf{Visual Understanding.} To validate the effectiveness of \abb in comprehensive visual understanding tasks, we apply it in several understanding and reasoning tasks, as some examples shown in Figure~\ref{fig:video_caption} and Figure~\ref{fig:vqa}. From the results, we can see the versatility of \abb in various tasks including visual captioning and visual question answering. Besides, our model has inherited some important capabilities from VILA \citep{lin2023vila} including multi-image understanding, in-context learning, as shown in Figure \ref{fig:icl} and Figure~\ref{fig:common_and_difference}. More visualizations can be found in the Appendix~\ref{appendix:2} and \ref{appendix:3}.

\begin{figure}[htbp]
    \centering
    \includegraphics[width=\textwidth]{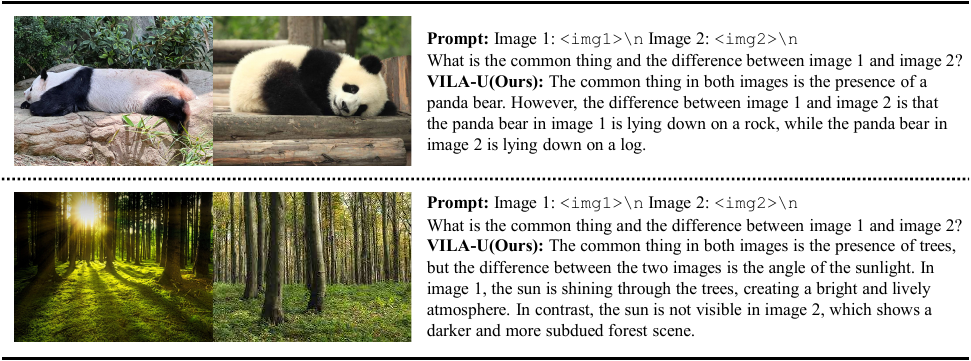}
    \caption{VILA-U can correctly reason over multiple images.}
    \label{fig:common_and_difference}
    \vspace{-10pt}
\end{figure}

\begin{figure}[t]
    \centering
    \includegraphics[width=\textwidth]{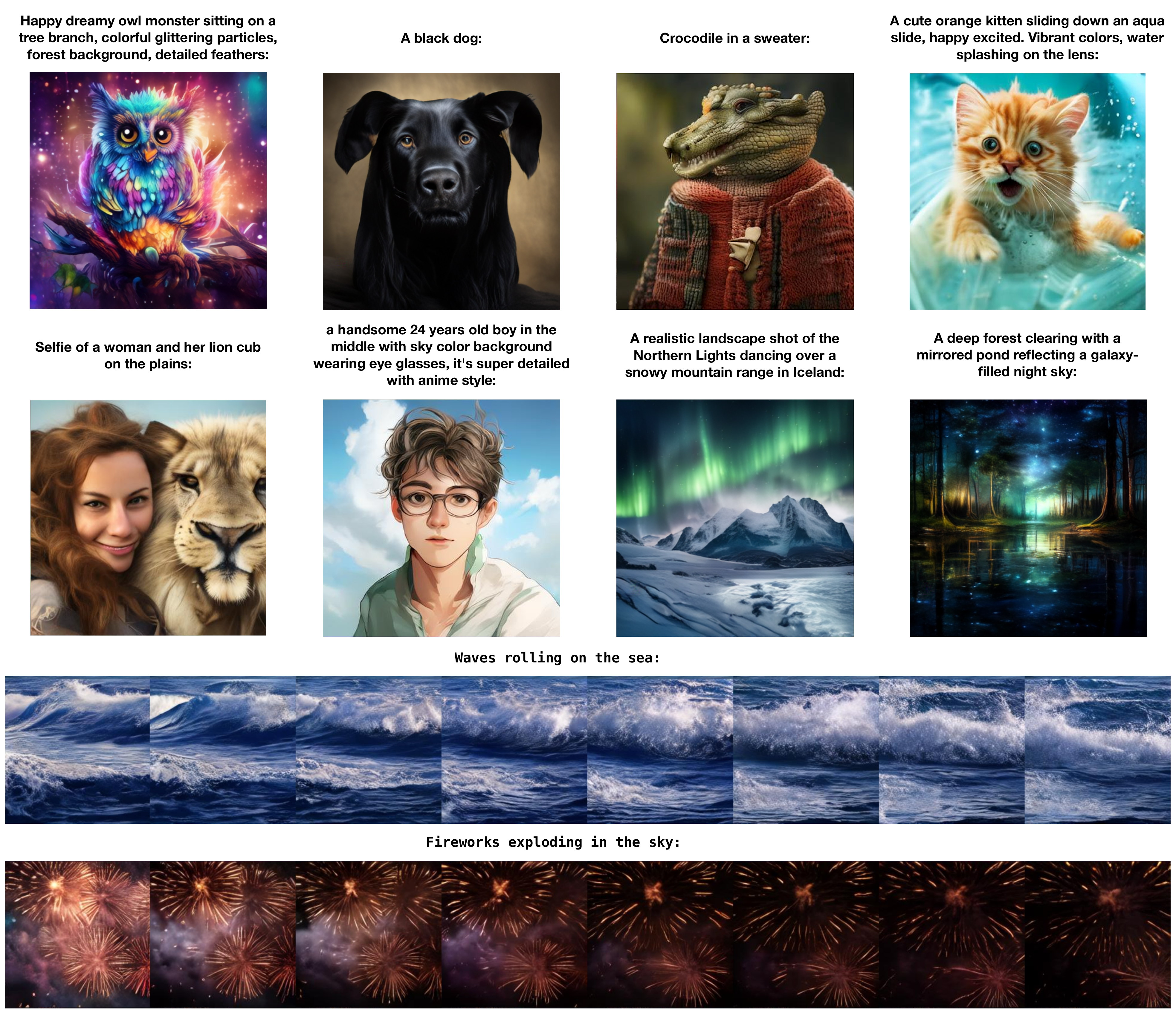}
    \caption{VILA-U can generate high-quality images and videos given text input.}
    \label{fig:generation}
    \vspace{-15pt}
\end{figure}

\textbf{Visual Generation.} We present some examples of the visual generation results in Figure \ref{fig:generation}. Our model can be employed in both image generation and video generation, even trained with a relatively small data corpus. In the given examples, our method can generate nice-looking images and continuous videos adhering to the user's input. More visualizations can be found in the appendix~\ref{appendix:4}.

\section{Ablation Study}
\label{sec:ablation}


\subsection{Impact of Contrastive Loss to Visual Understanding}

We include contrastive loss in vision tower training, which endows it with the text alignment ability. During our multi-modal training, such text alignment ability is crucial in enhancing modality fusion and performance on downstream visual language tasks. We validate the importance of this alignment by training the vision tower with and without the contrastive loss, evaluating its impact on visual language understanding performance. For this ablation study, we randomly sample 25M data from COYO-700M to train the vision tower. For multi-modal training, we use ShareGPT4V and MMC4 without text-image and text-video data. The results of the first two lines in Table \ref{tab:ablation_vision_tower} demonstrate the crucial role of text alignment in achieving strong visual language understanding performance. 
Scaling the dataset size from 25M to 700M further enhances performance, highlighting the importance of learning text alignment on a large-scale dataset.

\begin{table}[h]
\setlength{\tabcolsep}{5.5pt}
\centering
\caption{Impact of contrastive loss to visual understanding.}
\scalebox{0.8}{
\begin{tabular}{l c l | c | c c c c c c c c c}
\toprule
{\bf Pretrained Weights} & {\bf Data size} & {\bf Loss Type} & {\bf Top-1 Accuracy} & {\bf VQAv2}  &  {\bf POPE} & {\bf MME} & {\bf SEED} & {\bf MM-Vet}\\
\midrule
SigLIP-Large & 25M & Recon. & -- & 57.7 & 75.1 & 937.7 & 38.7 & 15.3 \\
SigLIP-Large & 25M & Recon. + Contra. &  62.9 & 68.0 & 83.7 & 1219 & 50.4 & 20.8\\
SigLIP-Large & 700M & Recon. + Contra. & 73.3  & 75.3 & 83.9 & 1336.2 & 56.3 & 27.7\\
\bottomrule
\end{tabular}
\vspace{-50pt}
}
\label{tab:ablation_vision_tower}
\end{table}

\vspace{-10pt}
\subsection{Impact of Contrastive Loss to Visual Generation}

We conduct two experiments to demonstrate the influence of contrastive loss to generation performance. For efficiency, we conduct only text-to-image pretraining and utilize Sheared-LLaMA-1.3B \citep{xia2023sheared} instead of LLaMA-2-7B as the LLM. In the first experiment, we use the RQ-VAE as the vision tower, which has an rFID of 1.30. In the second experiment, we employ our unified vision tower. Results are shown in Table \ref{tab:ablation_generation}. Our Unified Vision Tower yielded slightly worse FID results than the RQ-VAE on MJHQ-30K, possibly due to its inferior rFID resulting from the contrastive loss.

\begin{minipage}{0.65\textwidth}
\centering
\captionsetup{type=table}
\captionof{table}{Impact of contrastive loss to visual generation.}
\scalebox{0.8}{
\begin{tabular}{l l c c c}
\toprule
{\bf Vision Tower} & {\bf LLM} & {\bf Resolution} & {\bf rFID} $\downarrow$ & {\bf FID} $\downarrow$ \\
\midrule
RQ-VAE \citep{lee2022autoregressive} & Sheared-LLaMA-1.3B & 256 $\times$ 256 & 1.30 & 12.0 \\
Ours & Sheared-LLaMA-1.3B & 256 $\times$ 256 & 1.80 & 13.2 \\
\bottomrule
\end{tabular}
}
\label{tab:ablation_generation}
\end{minipage}
\hfill
\begin{minipage}{0.3\textwidth}
\centering
\captionsetup{type=table}
\captionof{table}{Impact of CFG.}
\scalebox{0.8}{
\begin{tabular}{c c}
\toprule
{\bf CFG Value} & {\bf FID} $\downarrow$ \\
\midrule
1.0 & 14.1 \\
2.0 & 13.0 \\
3.0 & 12.8 \\
5.0 & 13.2 \\

\bottomrule
\end{tabular}
}
\label{tab:ablation_cfg}
\end{minipage}

\subsection{Impact of Classifier-free Guidance}

We adopt classifier-free guidance during the visual content generation. We investigate the impact of the CFG value on our 256-resolution model. Results presented in Table \ref{tab:ablation_cfg} indicate that a CFG value of 3.0 yields the best FID score.

\section{Conclusion and Limitation}
\label{sec:conclusion}

We present \abbns, a novel and unified visual language model that integrates video, image and language understanding and generation tasks into one autoregressive next-token prediction framework. Our method is not only more concise than most VLMs that leverage additional components like diffusion models for unifying visual generation and understanding, but also demonstrates that autoregressive methods can achieve comparable performance to state-of-the-art VLMs. 
We believe \abb can serve as a general-purpose framework for diverse visual language tasks.

As demonstrated in Section 5.2, the introduction of contrastive loss impacts the reconstruction ability of the vision tower. Balancing these two capabilities within the unified vision tower presents an interesting and complex challenge that requires further exploration. Additionally, we currently do not observe significant synergy or mutual enhancement between understanding and generation tasks. In the future, we aim to investigate and explore more effective methods to enable these tasks to complement and reinforce each other, thereby fully realizing the untapped potential of a unified visual language model.

\bibliography{iclr2025_conference}

\begin{thebibliography}{66}
\providecommand{\natexlab}[1]{#1}
\providecommand{\url}[1]{\texttt{#1}}
\expandafter\ifx\csname urlstyle\endcsname\relax
  \providecommand{\doi}[1]{doi: #1}\else
  \providecommand{\doi}{doi: \begingroup \urlstyle{rm}\Url}\fi

\bibitem[Alayrac et~al.(2022)Alayrac, Donahue, Luc, Miech, Barr, Hasson, Lenc, Mensch, Millican, Reynolds, et~al.]{alayrac2022flamingo}
Jean-Baptiste Alayrac, Jeff Donahue, Pauline Luc, Antoine Miech, Iain Barr, Yana Hasson, Karel Lenc, Arthur Mensch, Katherine Millican, Malcolm Reynolds, et~al.
\newblock Flamingo: a visual language model for few-shot learning.
\newblock \emph{Advances in Neural Information Processing Systems}, 35:\penalty0 23716--23736, 2022.

\bibitem[Byeon et~al.(2022)Byeon, Park, Kim, Lee, Baek, and Kim]{coyo-700m}
Minwoo Byeon, Beomhee Park, Haecheon Kim, Sungjun Lee, Woonhyuk Baek, and Saehoon Kim.
\newblock Coyo-700m: Image-text pair dataset.
\newblock \url{https://github.com/kakaobrain/coyo-dataset}, 2022.

\bibitem[Caba~Heilbron et~al.(2015)Caba~Heilbron, Escorcia, Ghanem, and Carlos~Niebles]{activitynet}
Fabian Caba~Heilbron, Victor Escorcia, Bernard Ghanem, and Juan Carlos~Niebles.
\newblock Activitynet: A large-scale video benchmark for human activity understanding.
\newblock In \emph{Proceedings of the ieee conference on computer vision and pattern recognition}, pp.\  961--970, 2015.

\bibitem[Chen \& Dolan(2011)Chen and Dolan]{chen2011collecting}
David Chen and William~B Dolan.
\newblock Collecting highly parallel data for paraphrase evaluation.
\newblock In \emph{Proceedings of the 49th annual meeting of the association for computational linguistics: human language technologies}, pp.\  190--200, 2011.

\bibitem[Chen et~al.(2023)Chen, Li, Dong, Zhang, He, Wang, Zhao, and Lin]{sharegpt4v}
Lin Chen, Jisong Li, Xiaoyi Dong, Pan Zhang, Conghui He, Jiaqi Wang, Feng Zhao, and Dahua Lin.
\newblock Sharegpt4v: Improving large multi-modal models with better captions.
\newblock \emph{arXiv preprint arXiv:2311.12793}, 2023.

\bibitem[Chiang et~al.(2023)Chiang, Li, Lin, Sheng, Wu, Zhang, Zheng, Zhuang, Zhuang, Gonzalez, et~al.]{vicuna}
Wei-Lin Chiang, Zhuohan Li, Zi~Lin, Ying Sheng, Zhanghao Wu, Hao Zhang, Lianmin Zheng, Siyuan Zhuang, Yonghao Zhuang, Joseph~E Gonzalez, et~al.
\newblock Vicuna: An open-source chatbot impressing gpt-4 with 90\%* chatgpt quality.
\newblock \emph{See https://vicuna. lmsys. org (accessed 14 April 2023)}, 2\penalty0 (3):\penalty0 6, 2023.

\bibitem[Chung et~al.(2024)Chung, Hou, Longpre, Zoph, Tay, Fedus, Li, Wang, Dehghani, Brahma, et~al.]{chung2024scaling}
Hyung~Won Chung, Le~Hou, Shayne Longpre, Barret Zoph, Yi~Tay, William Fedus, Yunxuan Li, Xuezhi Wang, Mostafa Dehghani, Siddhartha Brahma, et~al.
\newblock Scaling instruction-finetuned language models.
\newblock \emph{Journal of Machine Learning Research}, 25\penalty0 (70):\penalty0 1--53, 2024.

\bibitem[Dai et~al.(2024)Dai, Li, Li, Tiong, Zhao, Wang, Li, Fung, and Hoi]{instructblip}
Wenliang Dai, Junnan Li, Dongxu Li, Anthony Meng~Huat Tiong, Junqi Zhao, Weisheng Wang, Boyang Li, Pascale~N Fung, and Steven Hoi.
\newblock Instructblip: Towards general-purpose vision-language models with instruction tuning.
\newblock \emph{Advances in Neural Information Processing Systems}, 36, 2024.

\bibitem[Deng et~al.(2009{\natexlab{a}})Deng, Dong, Socher, Li, Li, and Fei-Fei]{deng2009imagenet}
Jia Deng, Wei Dong, Richard Socher, Li-Jia Li, Kai Li, and Li~Fei-Fei.
\newblock Imagenet: A large-scale hierarchical image database.
\newblock In \emph{2009 IEEE conference on computer vision and pattern recognition}, pp.\  248--255. Ieee, 2009{\natexlab{a}}.

\bibitem[Deng et~al.(2009{\natexlab{b}})Deng, Dong, Socher, Li, Li, and Fei-Fei]{imagenet}
Jia Deng, Wei Dong, Richard Socher, Li-Jia Li, Kai Li, and Li~Fei-Fei.
\newblock Imagenet: A large-scale hierarchical image database.
\newblock In \emph{2009 IEEE conference on computer vision and pattern recognition}, pp.\  248--255. Ieee, 2009{\natexlab{b}}.

\bibitem[Esser et~al.(2021)Esser, Rombach, and Ommer]{VQGAN}
Patrick Esser, Robin Rombach, and Bjorn Ommer.
\newblock Taming transformers for high-resolution image synthesis.
\newblock In \emph{Proceedings of the IEEE/CVF conference on computer vision and pattern recognition}, pp.\  12873--12883, 2021.

\bibitem[Fu et~al.(2024)Fu, Chen, Shen, Qin, Zhang, Lin, Yang, Zheng, Li, Sun, Wu, and Ji]{mme}
Chaoyou Fu, Peixian Chen, Yunhang Shen, Yulei Qin, Mengdan Zhang, Xu~Lin, Jinrui Yang, Xiawu Zheng, Ke~Li, Xing Sun, Yunsheng Wu, and Rongrong Ji.
\newblock Mme: A comprehensive evaluation benchmark for multimodal large language models, 2024.

\bibitem[Ge et~al.(2023{\natexlab{a}})Ge, Ge, Zeng, Wang, and Shan]{ge2023planting}
Yuying Ge, Yixiao Ge, Ziyun Zeng, Xintao Wang, and Ying Shan.
\newblock Planting a seed of vision in large language model.
\newblock \emph{arXiv preprint arXiv:2307.08041}, 2023{\natexlab{a}}.

\bibitem[Ge et~al.(2023{\natexlab{b}})Ge, Zhao, Zeng, Ge, Li, Wang, and Shan]{seed-llama}
Yuying Ge, Sijie Zhao, Ziyun Zeng, Yixiao Ge, Chen Li, Xintao Wang, and Ying Shan.
\newblock Making llama see and draw with seed tokenizer.
\newblock In \emph{The Twelfth International Conference on Learning Representations}, 2023{\natexlab{b}}.

\bibitem[Ge et~al.(2024)Ge, Zhao, Zhu, Ge, Yi, Song, Li, Ding, and Shan]{seed-x}
Yuying Ge, Sijie Zhao, Jinguo Zhu, Yixiao Ge, Kun Yi, Lin Song, Chen Li, Xiaohan Ding, and Ying Shan.
\newblock Seed-x: Multimodal models with unified multi-granularity comprehension and generation.
\newblock \emph{arXiv preprint arXiv:2404.14396}, 2024.

\bibitem[Goyal et~al.(2017)Goyal, Khot, Summers{-}Stay, Batra, and Parikh]{vqa_v2}
Yash Goyal, Tejas Khot, Douglas Summers{-}Stay, Dhruv Batra, and Devi Parikh.
\newblock Making the {V} in {VQA} matter: Elevating the role of image understanding in {V}isual {Q}uestion {A}nswering.
\newblock In \emph{Conference on Computer Vision and Pattern Recognition (CVPR)}, 2017.

\bibitem[Ho \& Salimans(2022)Ho and Salimans]{cfg}
Jonathan Ho and Tim Salimans.
\newblock Classifier-free diffusion guidance.
\newblock \emph{arXiv preprint arXiv:2207.12598}, 2022.

\bibitem[Hong et~al.(2022)Hong, Ding, Zheng, Liu, and Tang]{hong2022cogvideo}
Wenyi Hong, Ming Ding, Wendi Zheng, Xinghan Liu, and Jie Tang.
\newblock Cogvideo: Large-scale pretraining for text-to-video generation via transformers.
\newblock \emph{arXiv preprint arXiv:2205.15868}, 2022.

\bibitem[Huang et~al.(2024)Huang, He, Yu, Zhang, Si, Jiang, Zhang, Wu, Jin, Chanpaisit, et~al.]{vbench}
Ziqi Huang, Yinan He, Jiashuo Yu, Fan Zhang, Chenyang Si, Yuming Jiang, Yuanhan Zhang, Tianxing Wu, Qingyang Jin, Nattapol Chanpaisit, et~al.
\newblock Vbench: Comprehensive benchmark suite for video generative models.
\newblock In \emph{Proceedings of the IEEE/CVF Conference on Computer Vision and Pattern Recognition}, pp.\  21807--21818, 2024.

\bibitem[Hudson \& Manning(2019)Hudson and Manning]{gqa}
Drew~A Hudson and Christopher~D Manning.
\newblock Gqa: A new dataset for real-world visual reasoning and compositional question answering.
\newblock In \emph{Proceedings of the IEEE/CVF conference on computer vision and pattern recognition}, pp.\  6700--6709, 2019.

\bibitem[Jiang et~al.(2024)Jiang, Sablayrolles, Roux, Mensch, Savary, Bamford, Chaplot, Casas, Hanna, Bressand, et~al.]{mixtral}
Albert~Q Jiang, Alexandre Sablayrolles, Antoine Roux, Arthur Mensch, Blanche Savary, Chris Bamford, Devendra~Singh Chaplot, Diego de~las Casas, Emma~Bou Hanna, Florian Bressand, et~al.
\newblock Mixtral of experts.
\newblock \emph{arXiv:2401.04088}, 2024.

\bibitem[Jin et~al.(2023)Jin, Xu, Chen, Liao, Tan, Huang, Bin, Song, ZHANG, Ou, et~al.]{jin2023unified}
Yang Jin, Kun Xu, Liwei Chen, Chao Liao, Jianchao Tan, Quzhe Huang, CHEN Bin, Chengru Song, Di~ZHANG, Wenwu Ou, et~al.
\newblock Unified language-vision pretraining in llm with dynamic discrete visual tokenization.
\newblock In \emph{The Twelfth International Conference on Learning Representations}, 2023.

\bibitem[Jin et~al.(2024)Jin, Sun, Xu, Chen, Jiang, Huang, Song, Liu, Zhang, Song, et~al.]{video-lavit}
Yang Jin, Zhicheng Sun, Kun Xu, Liwei Chen, Hao Jiang, Quzhe Huang, Chengru Song, Yuliang Liu, Di~Zhang, Yang Song, et~al.
\newblock Video-lavit: Unified video-language pre-training with decoupled visual-motional tokenization.
\newblock \emph{arXiv preprint arXiv:2402.03161}, 2024.

\bibitem[Lee et~al.(2022)Lee, Kim, Kim, Cho, and Han]{lee2022autoregressive}
Doyup Lee, Chiheon Kim, Saehoon Kim, Minsu Cho, and Wook-Shin Han.
\newblock Autoregressive image generation using residual quantization.
\newblock In \emph{Proceedings of the IEEE/CVF Conference on Computer Vision and Pattern Recognition}, pp.\  11523--11532, 2022.

\bibitem[Li et~al.(2023{\natexlab{a}})Li, Wang, Wang, Ge, Ge, and Shan]{seed}
Bohao Li, Rui Wang, Guangzhi Wang, Yuying Ge, Yixiao Ge, and Ying Shan.
\newblock Seed-bench: Benchmarking multimodal llms with generative comprehension.
\newblock \emph{arXiv preprint arXiv:2307.16125}, 2023{\natexlab{a}}.

\bibitem[Li et~al.(2024)Li, Kamko, Akhgari, Sabet, Xu, and Doshi]{playgroundv2.5}
Daiqing Li, Aleks Kamko, Ehsan Akhgari, Ali Sabet, Linmiao Xu, and Suhail Doshi.
\newblock Playground v2. 5: Three insights towards enhancing aesthetic quality in text-to-image generation.
\newblock \emph{arXiv preprint arXiv:2402.17245}, 2024.

\bibitem[Li et~al.(2022)Li, Li, Xiong, and Hoi]{blip}
Junnan Li, Dongxu Li, Caiming Xiong, and Steven Hoi.
\newblock Blip: Bootstrapping language-image pre-training for unified vision-language understanding and generation.
\newblock In \emph{ICML}, 2022.

\bibitem[Li et~al.(2023{\natexlab{b}})Li, Li, Savarese, and Hoi]{blip2}
Junnan Li, Dongxu Li, Silvio Savarese, and Steven Hoi.
\newblock Blip-2: Bootstrapping language-image pre-training with frozen image encoders and large language models.
\newblock In \emph{International conference on machine learning}, pp.\  19730--19742. PMLR, 2023{\natexlab{b}}.

\bibitem[Li et~al.(2023{\natexlab{c}})Li, Zhang, Wang, Zhong, Chen, Chu, Liu, and Jia]{li2024mgm}
Yanwei Li, Yuechen Zhang, Chengyao Wang, Zhisheng Zhong, Yixin Chen, Ruihang Chu, Shaoteng Liu, and Jiaya Jia.
\newblock Mini-gemini: Mining the potential of multi-modality vision language models.
\newblock \emph{arXiv:2403.18814}, 2023{\natexlab{c}}.

\bibitem[Li et~al.(2023{\natexlab{d}})Li, Du, Zhou, Wang, Zhao, and Wen]{POPE}
Yifan Li, Yifan Du, Kun Zhou, Jinpeng Wang, Wayne~Xin Zhao, and Ji-Rong Wen.
\newblock Evaluating object hallucination in large vision-language models.
\newblock \emph{arXiv preprint arXiv:2305.10355}, 2023{\natexlab{d}}.

\bibitem[Li et~al.(2016)Li, Song, Cao, Tetreault, Goldberg, Jaimes, and Luo]{tgif}
Yuncheng Li, Yale Song, Liangliang Cao, Joel Tetreault, Larry Goldberg, Alejandro Jaimes, and Jiebo Luo.
\newblock Tgif: A new dataset and benchmark on animated gif description.
\newblock In \emph{Proceedings of the IEEE Conference on Computer Vision and Pattern Recognition}, pp.\  4641--4650, 2016.

\bibitem[Lin et~al.(2023)Lin, Yin, Ping, Lu, Molchanov, Tao, Mao, Kautz, Shoeybi, and Han]{lin2023vila}
Ji~Lin, Hongxu Yin, Wei Ping, Yao Lu, Pavlo Molchanov, Andrew Tao, Huizi Mao, Jan Kautz, Mohammad Shoeybi, and Song Han.
\newblock Vila: On pre-training for visual language models, 2023.

\bibitem[Lin et~al.(2024)Lin, Pathak, Li, Li, Xia, Neubig, Zhang, and Ramanan]{GenAI-Bench}
Zhiqiu Lin, Deepak Pathak, Baiqi Li, Jiayao Li, Xide Xia, Graham Neubig, Pengchuan Zhang, and Deva Ramanan.
\newblock Evaluating text-to-visual generation with image-to-text generation.
\newblock \emph{arXiv preprint arXiv:2404.01291}, 2024.

\bibitem[Liu et~al.(2024{\natexlab{a}})Liu, Yan, Zaharia, and Abbeel]{lwm}
Hao Liu, Wilson Yan, Matei Zaharia, and Pieter Abbeel.
\newblock World model on million-length video and language with ringattention.
\newblock \emph{arXiv preprint arXiv:2402.08268}, 2024{\natexlab{a}}.

\bibitem[Liu et~al.(2024{\natexlab{b}})Liu, Li, Wu, and Lee]{llava}
Haotian Liu, Chunyuan Li, Qingyang Wu, and Yong~Jae Lee.
\newblock Visual instruction tuning.
\newblock \emph{Advances in neural information processing systems}, 36, 2024{\natexlab{b}}.

\bibitem[Lu et~al.(2023)Lu, Clark, Lee, Zhang, Khosla, Marten, Hoiem, and Kembhavi]{Unified-io2}
Jiasen Lu, Christopher Clark, Sangho Lee, Zichen Zhang, Savya Khosla, Ryan Marten, Derek Hoiem, and Aniruddha Kembhavi.
\newblock Unified-io 2: Scaling autoregressive multimodal models with vision, language, audio, and action.
\newblock \emph{arXiv preprint arXiv:2312.17172}, 2023.

\bibitem[Luo et~al.(2024)Luo, Li, Chen, He, Lin, Liu, Zhang, Song, Xia, Liu, et~al.]{luo2024deem}
Run Luo, Yunshui Li, Longze Chen, Wanwei He, Ting-En Lin, Ziqiang Liu, Lei Zhang, Zikai Song, Xiaobo Xia, Tongliang Liu, et~al.
\newblock Deem: Diffusion models serve as the eyes of large language models for image perception.
\newblock \emph{arXiv preprint arXiv:2405.15232}, 2024.

\bibitem[Nan et~al.(2024)Nan, Xie, Zhou, Fan, Yang, Chen, Li, Yang, and Tai]{nan2024openvid}
Kepan Nan, Rui Xie, Penghao Zhou, Tiehan Fan, Zhenheng Yang, Zhijie Chen, Xiang Li, Jian Yang, and Ying Tai.
\newblock Openvid-1m: A large-scale high-quality dataset for text-to-video generation.
\newblock \emph{arXiv preprint arXiv:2407.02371}, 2024.

\bibitem[OpenAI(2023)]{ChatGPT}
OpenAI.
\newblock Chatgpt.
\newblock \url{https://openai.com/blog/chatgpt/}, 2023.

\bibitem[Ouyang et~al.(2022)Ouyang, Wu, Jiang, Almeida, Wainwright, Mishkin, Zhang, Agarwal, Slama, Ray, et~al.]{ouyang2022training}
Long Ouyang, Jeffrey Wu, Xu~Jiang, Diogo Almeida, Carroll Wainwright, Pamela Mishkin, Chong Zhang, Sandhini Agarwal, Katarina Slama, Alex Ray, et~al.
\newblock Training language models to follow instructions with human feedback.
\newblock \emph{Advances in neural information processing systems}, 35:\penalty0 27730--27744, 2022.

\bibitem[Podell et~al.(2023)Podell, English, Lacey, Blattmann, Dockhorn, M{\"u}ller, Penna, and Rombach]{sdxl}
Dustin Podell, Zion English, Kyle Lacey, Andreas Blattmann, Tim Dockhorn, Jonas M{\"u}ller, Joe Penna, and Robin Rombach.
\newblock Sdxl: Improving latent diffusion models for high-resolution image synthesis.
\newblock \emph{arXiv preprint arXiv:2307.01952}, 2023.

\bibitem[Radford et~al.(2021)Radford, Kim, Hallacy, Ramesh, Goh, Agarwal, Sastry, Askell, Mishkin, Clark, et~al.]{clip}
Alec Radford, Jong~Wook Kim, Chris Hallacy, Aditya Ramesh, Gabriel Goh, Sandhini Agarwal, Girish Sastry, Amanda Askell, Pamela Mishkin, Jack Clark, et~al.
\newblock Learning transferable visual models from natural language supervision.
\newblock In \emph{International conference on machine learning}, pp.\  8748--8763. PMLR, 2021.

\bibitem[Rombach et~al.(2022{\natexlab{a}})Rombach, Blattmann, Lorenz, Esser, and Ommer]{ldm}
Robin Rombach, Andreas Blattmann, Dominik Lorenz, Patrick Esser, and Bj{\"o}rn Ommer.
\newblock High-resolution image synthesis with latent diffusion models.
\newblock In \emph{Proceedings of the IEEE/CVF conference on computer vision and pattern recognition}, pp.\  10684--10695, 2022{\natexlab{a}}.

\bibitem[Rombach et~al.(2022{\natexlab{b}})Rombach, Blattmann, Lorenz, Esser, and Ommer]{rombach2022high}
Robin Rombach, Andreas Blattmann, Dominik Lorenz, Patrick Esser, and Bj{\"o}rn Ommer.
\newblock High-resolution image synthesis with latent diffusion models.
\newblock In \emph{Proceedings of the IEEE/CVF conference on computer vision and pattern recognition}, pp.\  10684--10695, 2022{\natexlab{b}}.

\bibitem[Singh et~al.(2019)Singh, Natarajan, Shah, Jiang, Chen, Batra, Parikh, and Rohrbach]{textvqa}
Amanpreet Singh, Vivek Natarajan, Meet Shah, Yu~Jiang, Xinlei Chen, Dhruv Batra, Devi Parikh, and Marcus Rohrbach.
\newblock Towards vqa models that can read.
\newblock In \emph{Proceedings of the IEEE/CVF conference on computer vision and pattern recognition}, pp.\  8317--8326, 2019.

\bibitem[Sun et~al.(2024)Sun, Jiang, Chen, Zhang, Peng, Luo, and Yuan]{llamagen}
Peize Sun, Yi~Jiang, Shoufa Chen, Shilong Zhang, Bingyue Peng, Ping Luo, and Zehuan Yuan.
\newblock Autoregressive model beats diffusion: Llama for scalable image generation, 2024.
\newblock URL \url{https://arxiv.org/abs/2406.06525}.

\bibitem[Sun et~al.(2023{\natexlab{a}})Sun, Cui, Zhang, Zhang, Yu, Luo, Wang, Rao, Liu, Huang, and Wang]{Emu2}
Quan Sun, Yufeng Cui, Xiaosong Zhang, Fan Zhang, Qiying Yu, Zhengxiong Luo, Yueze Wang, Yongming Rao, Jingjing Liu, Tiejun Huang, and Xinlong Wang.
\newblock Generative multimodal models are in-context learners.
\newblock \emph{arXiv preprint arXiv:2312.13286}, 2023{\natexlab{a}}.

\bibitem[Sun et~al.(2023{\natexlab{b}})Sun, Yu, Cui, Zhang, Zhang, Wang, Gao, Liu, Huang, and Wang]{Emu}
Quan Sun, Qiying Yu, Yufeng Cui, Fan Zhang, Xiaosong Zhang, Yueze Wang, Hongcheng Gao, Jingjing Liu, Tiejun Huang, and Xinlong Wang.
\newblock Generative pretraining in multimodality.
\newblock \emph{arXiv preprint arXiv:2307.05222}, 2023{\natexlab{b}}.

\bibitem[Team(2024)]{chameleon}
Chameleon Team.
\newblock Chameleon: Mixed-modal early-fusion foundation models, 2024.

\bibitem[Tian et~al.(2024{\natexlab{a}})Tian, Zhu, Xiong, Wang, Chen, Wang, Chen, Lu, Lu, Zhou, et~al.]{tian2024mm}
Changyao Tian, Xizhou Zhu, Yuwen Xiong, Weiyun Wang, Zhe Chen, Wenhai Wang, Yuntao Chen, Lewei Lu, Tong Lu, Jie Zhou, et~al.
\newblock Mm-interleaved: Interleaved image-text generative modeling via multi-modal feature synchronizer.
\newblock \emph{arXiv preprint arXiv:2401.10208}, 2024{\natexlab{a}}.

\bibitem[Tian et~al.(2024{\natexlab{b}})Tian, Jiang, Yuan, Peng, and Wang]{var}
Keyu Tian, Yi~Jiang, Zehuan Yuan, Bingyue Peng, and Liwei Wang.
\newblock Visual autoregressive modeling: Scalable image generation via next-scale prediction, 2024{\natexlab{b}}.
\newblock URL \url{https://arxiv.org/abs/2404.02905}.

\bibitem[Touvron et~al.(2023{\natexlab{a}})Touvron, Lavril, Izacard, Martinet, Lachaux, Lacroix, Rozi{\`e}re, Goyal, Hambro, Azhar, Rodriguez, Joulin, Grave, and Lample]{llama}
Hugo Touvron, Thibaut Lavril, Gautier Izacard, Xavier Martinet, Marie-Anne Lachaux, Timoth{\'e}e Lacroix, Baptiste Rozi{\`e}re, Naman Goyal, Eric Hambro, Faisal Azhar, Aurelien Rodriguez, Armand Joulin, Edouard Grave, and Guillaume Lample.
\newblock Llama: Open and efficient foundation language models.
\newblock \emph{arXiv:2302.13971}, 2023{\natexlab{a}}.

\bibitem[Touvron et~al.(2023{\natexlab{b}})Touvron, Lavril, Izacard, Martinet, Lachaux, Lacroix, Rozi{\`e}re, Goyal, Hambro, Azhar, et~al.]{touvron2023llama}
Hugo Touvron, Thibaut Lavril, Gautier Izacard, Xavier Martinet, Marie-Anne Lachaux, Timoth{\'e}e Lacroix, Baptiste Rozi{\`e}re, Naman Goyal, Eric Hambro, Faisal Azhar, et~al.
\newblock Llama: Open and efficient foundation language models.
\newblock \emph{arXiv preprint arXiv:2302.13971}, 2023{\natexlab{b}}.

\bibitem[Vaswani et~al.(2017)Vaswani, Shazeer, Parmar, Uszkoreit, Jones, Gomez, Kaiser, and Polosukhin]{NIPS2017Transformer}
Ashish Vaswani, Noam Shazeer, Niki Parmar, Jakob Uszkoreit, Llion Jones, Aidan~N Gomez, \L~ukasz Kaiser, and Illia Polosukhin.
\newblock Attention is all you need.
\newblock In I.~Guyon, U.~Von Luxburg, S.~Bengio, H.~Wallach, R.~Fergus, S.~Vishwanathan, and R.~Garnett (eds.), \emph{Advances in Neural Information Processing Systems}, volume~30. Curran Associates, Inc., 2017.
\newblock URL \url{https://proceedings.neurips.cc/paper_files/paper/2017/file/3f5ee243547dee91fbd053c1c4a845aa-Paper.pdf}.

\bibitem[Xia et~al.(2023)Xia, Gao, Zeng, and Chen]{xia2023sheared}
Mengzhou Xia, Tianyu Gao, Zhiyuan Zeng, and Danqi Chen.
\newblock Sheared llama: Accelerating language model pre-training via structured pruning.
\newblock \emph{arXiv preprint arXiv:2310.06694}, 2023.

\bibitem[Xie et~al.(2024)Xie, Mao, Bai, Zhang, Wang, Lin, Gu, Chen, Yang, and Shou]{xie2024show}
Jinheng Xie, Weijia Mao, Zechen Bai, David~Junhao Zhang, Weihao Wang, Kevin~Qinghong Lin, Yuchao Gu, Zhijie Chen, Zhenheng Yang, and Mike~Zheng Shou.
\newblock Show-o: One single transformer to unify multimodal understanding and generation.
\newblock \emph{arXiv preprint arXiv:2408.12528}, 2024.

\bibitem[Xu et~al.(2017)Xu, Zhao, Xiao, Wu, Zhang, He, and Zhuang]{MSVD}
Dejing Xu, Zhou Zhao, Jun Xiao, Fei Wu, Hanwang Zhang, Xiangnan He, and Yueting Zhuang.
\newblock Video question answering via gradually refined attention over appearance and motion.
\newblock In \emph{Proceedings of the 25th ACM international conference on Multimedia}, pp.\  1645--1653, 2017.

\bibitem[Yang et~al.(2024)Yang, Teng, Zheng, Ding, Huang, Xu, Yang, Hong, Zhang, Feng, et~al.]{yang2024cogvideox}
Zhuoyi Yang, Jiayan Teng, Wendi Zheng, Ming Ding, Shiyu Huang, Jiazheng Xu, Yuanming Yang, Wenyi Hong, Xiaohan Zhang, Guanyu Feng, et~al.
\newblock Cogvideox: Text-to-video diffusion models with an expert transformer.
\newblock \emph{arXiv preprint arXiv:2408.06072}, 2024.

\bibitem[Yu et~al.(2021)Yu, Li, Koh, Zhang, Pang, Qin, Ku, Xu, Baldridge, and Wu]{vit-vqgan}
Jiahui Yu, Xin Li, Jing~Yu Koh, Han Zhang, Ruoming Pang, James Qin, Alexander Ku, Yuanzhong Xu, Jason Baldridge, and Yonghui Wu.
\newblock Vector-quantized image modeling with improved vqgan.
\newblock \emph{arXiv preprint arXiv:2110.04627}, 2021.

\bibitem[Yu et~al.(2022)Yu, Wang, Vasudevan, Yeung, Seyedhosseini, and Wu]{CoCa}
Jiahui Yu, Zirui Wang, Vijay Vasudevan, Legg Yeung, Mojtaba Seyedhosseini, and Yonghui Wu.
\newblock Coca: Contrastive captioners are image-text foundation models, 2022.

\bibitem[Yu et~al.(2023{\natexlab{a}})Yu, Shi, Pasunuru, Muller, Golovneva, Wang, Babu, Tang, Karrer, Sheynin, et~al.]{CM3Leon}
Lili Yu, Bowen Shi, Ramakanth Pasunuru, Benjamin Muller, Olga Golovneva, Tianlu Wang, Arun Babu, Binh Tang, Brian Karrer, Shelly Sheynin, et~al.
\newblock Scaling autoregressive multi-modal models: Pretraining and instruction tuning.
\newblock \emph{arXiv preprint arXiv:2309.02591}, 2\penalty0 (3), 2023{\natexlab{a}}.

\bibitem[Yu et~al.(2023{\natexlab{b}})Yu, Yang, Li, Wang, Lin, Liu, Wang, and Wang]{mm-vet}
Weihao Yu, Zhengyuan Yang, Linjie Li, Jianfeng Wang, Kevin Lin, Zicheng Liu, Xinchao Wang, and Lijuan Wang.
\newblock Mm-vet: Evaluating large multimodal models for integrated capabilities.
\newblock \emph{arXiv preprint arXiv:2308.02490}, 2023{\natexlab{b}}.

\bibitem[Zhai et~al.(2023)Zhai, Mustafa, Kolesnikov, and Beyer]{zhai2023sigmoid}
Xiaohua Zhai, Basil Mustafa, Alexander Kolesnikov, and Lucas Beyer.
\newblock Sigmoid loss for language image pre-training.
\newblock In \emph{Proceedings of the IEEE/CVF International Conference on Computer Vision}, pp.\  11975--11986, 2023.

\bibitem[Zhan et~al.(2024)Zhan, Dai, Ye, Zhou, Zhang, Liu, Zhang, Yuan, Zhang, Li, et~al.]{zhan2024anygpt}
Jun Zhan, Junqi Dai, Jiasheng Ye, Yunhua Zhou, Dong Zhang, Zhigeng Liu, Xin Zhang, Ruibin Yuan, Ge~Zhang, Linyang Li, et~al.
\newblock Anygpt: Unified multimodal llm with discrete sequence modeling.
\newblock \emph{arXiv preprint arXiv:2402.12226}, 2024.

\bibitem[Zheng et~al.()Zheng, Peng, Yang, Shen, Li, Liu, Zhou, Li, and You]{zheng1open}
Zangwei Zheng, Xiangyu Peng, Tianji Yang, Chenhui Shen, Shenggui Li, Hongxin Liu, Yukun Zhou, Tianyi Li, and Yang You.
\newblock Open-sora: Democratizing efficient video production for all, march 2024.
\newblock \emph{URL https://github. com/hpcaitech/Open-Sora}, 1\penalty0 (3):\penalty0 4.

\bibitem[Zhu et~al.(2024)Zhu, Hessel, Awadalla, Gadre, Dodge, Fang, Yu, Schmidt, Wang, and Choi]{mmc4}
Wanrong Zhu, Jack Hessel, Anas Awadalla, Samir~Yitzhak Gadre, Jesse Dodge, Alex Fang, Youngjae Yu, Ludwig Schmidt, William~Yang Wang, and Yejin Choi.
\newblock Multimodal c4: An open, billion-scale corpus of images interleaved with text.
\newblock \emph{Advances in Neural Information Processing Systems}, 36, 2024.

\end{thebibliography}
\bibliographystyle{iclr2025_conference}

\newpage
\appendix
\appendix
\section*{Appendix}

\section{Difference with related works}
\label{appendix:A}

Prior to VILA-U, unified visual language models were dominated by two mainstream approaches:

(1) Represented by LWM, CM3Leon and Show-o which utilizes a VQGAN-based tokenizer to convert visual inputs into discrete tokens. However, as these tokenizers are trained solely with a reconstruction objective, the resulting tokens lack rich semantic information. This limitation leads to poor performance on multimodal understanding tasks. But it can easily support autoregressive visual generation and the generated visual tokens can be seamlessly decoded into visual outputs using the lightweight decoder of VQGAN.

(2) Represented by AnyGPT SEED-LLaMa and LaViT, which utilizes a codebook to quantize features produced by a pre-trained ViT model like CLIP. Since CLIP features encode rich semantic information, these approaches generally achieve significantly better performance on understanding tasks compared to VQGAN-based tokenizers. However, these tokenizers lack decoding capability, requiring an external visual generation model, such as a diffusion model, to use the generated visual tokens as conditions for producing visual outputs.

Compared to these two mainstream approaches, VILA-U introduces a solution that addresses the limitations of both. We design a unified vision tower that extracts features with rich semantic information, similar to CLIP, while also supporting image reconstruction capabilities akin to VQGAN. This is achieved by incorporating both reconstruction loss and contrastive loss into the autoencoder training process, along with utilizing residual quantization to enhance the representation capability of the visual features. Building on this foundation, we develop a single end-to-end autoregressive framework that eliminates the need for external visual generation models required by approach 2 and significantly outperforms the understanding results of methods in approach 1.

\section{Qualitative Results}
\subsection{Reconstruction}
\label{appendix:1}

\begin{figure}[htbp]
  \centering
  \includegraphics[width=0.95\textwidth]{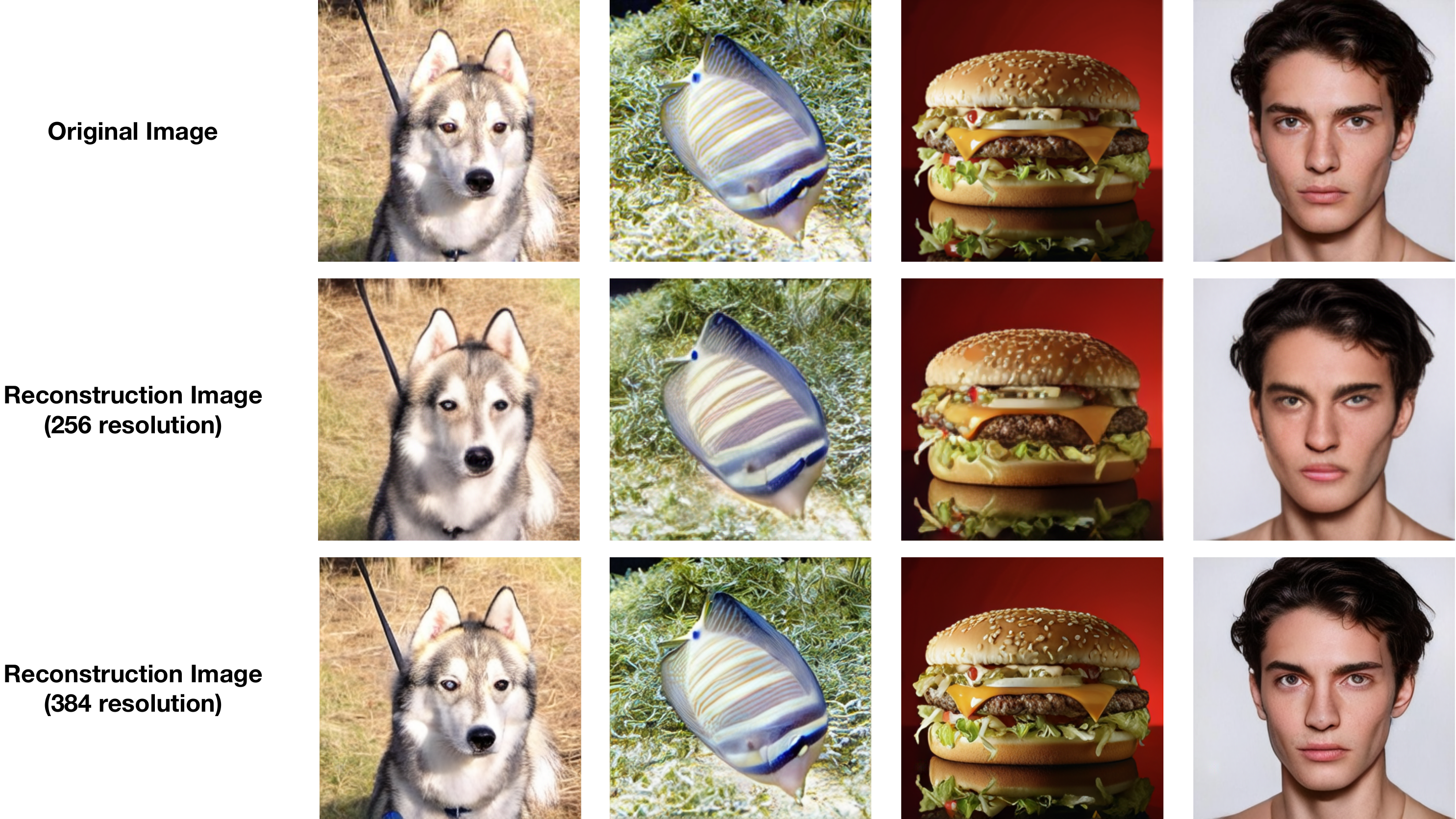}
  \caption{Visualization of the reconstruction results from text-aligned discrete visual tokens.}
  \label{fig:reconstruction}
\end{figure}

We present qualitative reconstruction results in Figure \ref{fig:reconstruction} for our 256 / 384 resolution vision tower. These vision towers effectively reconstruct images in detail using text-aligned discrete visual tokens.

\subsection{Visual Understanding}
\label{appendix:2}

\begin{figure}[h]
    \centering
    \includegraphics[width=0.56\textwidth]{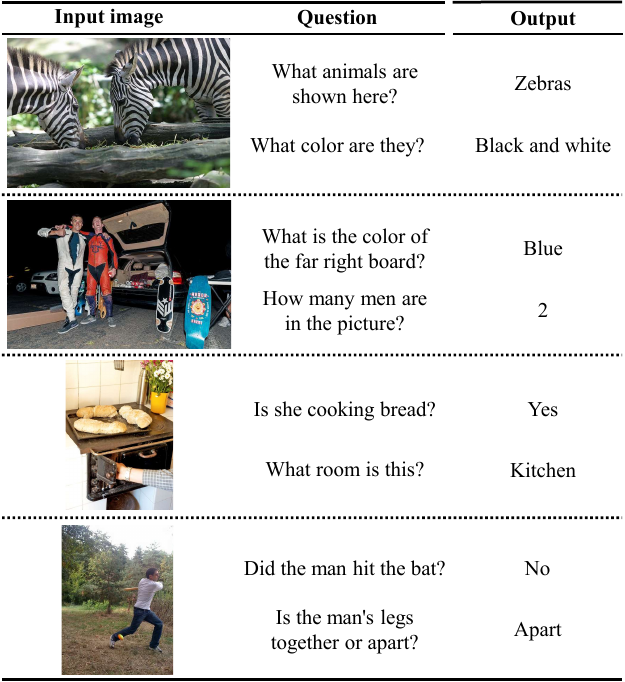}
    \caption{Image understanding results. Examples are taken from the test split of VQAv2 dataset.}
    \label{fig:vqa_more}
\end{figure}

\begin{figure}[htbp]
    \centering
    \includegraphics[width=\textwidth]{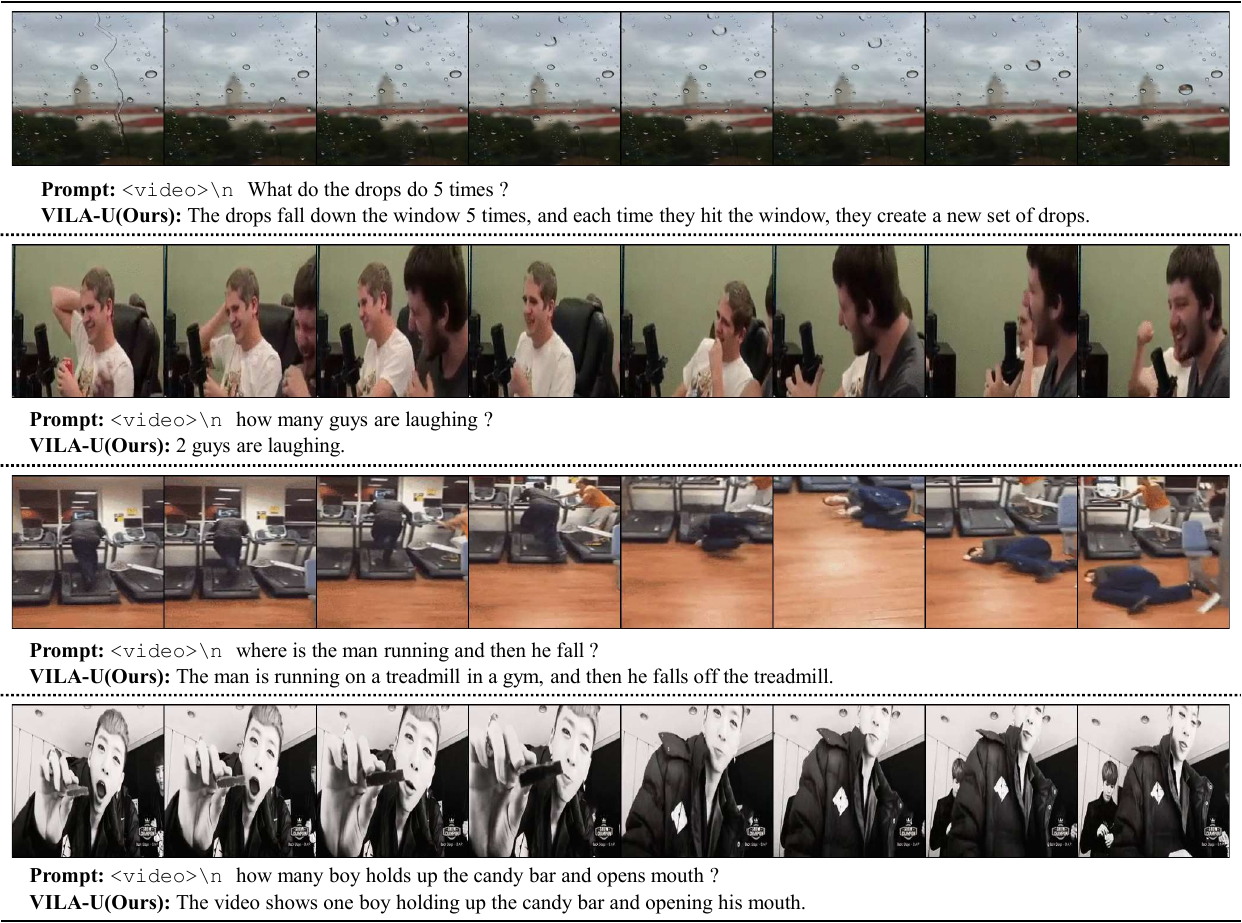}
    \caption{Video understanding results. Examples are taken from the test split of TGIF dataset.}
    \label{fig:tgif}
\end{figure}

We provide more image understanding and video understanding examples in Figure~\ref{fig:vqa_more} and Figure~\ref{fig:tgif}. VILA-U successfully answers the questions accurately.

\subsection{In-context Learning Examples}
\label{appendix:3}

\begin{figure}[htbp]
    \centering
    \includegraphics[width=\textwidth]{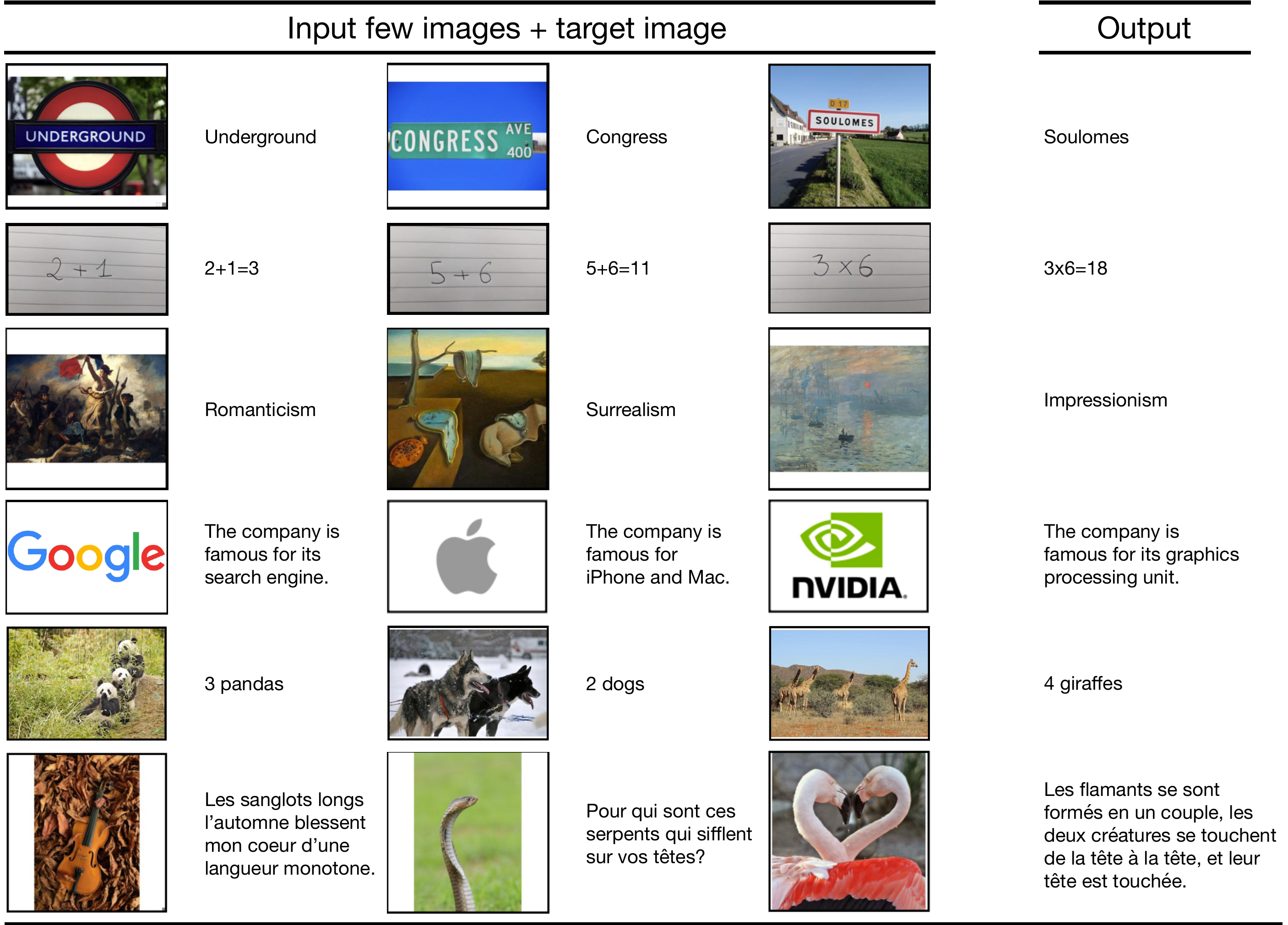}
    \caption{In-context learning examples. We try all in-context learning examples in \citet{lin2023vila}. The results demonstrate that VILA-U has inherited good in-context learning capabilties.}
    \label{fig:icl_more}
\end{figure}

We provide more qualitative results to demonstrate in-context learning capabilities of VILA-U in Figure~\ref{fig:icl_more}. VILA-U exhibits good in-context learning capabilties.

\subsection{Visual Generation}
\label{appendix:4}

\begin{figure}[htbp]
    \centering
    \includegraphics[width=\textwidth]{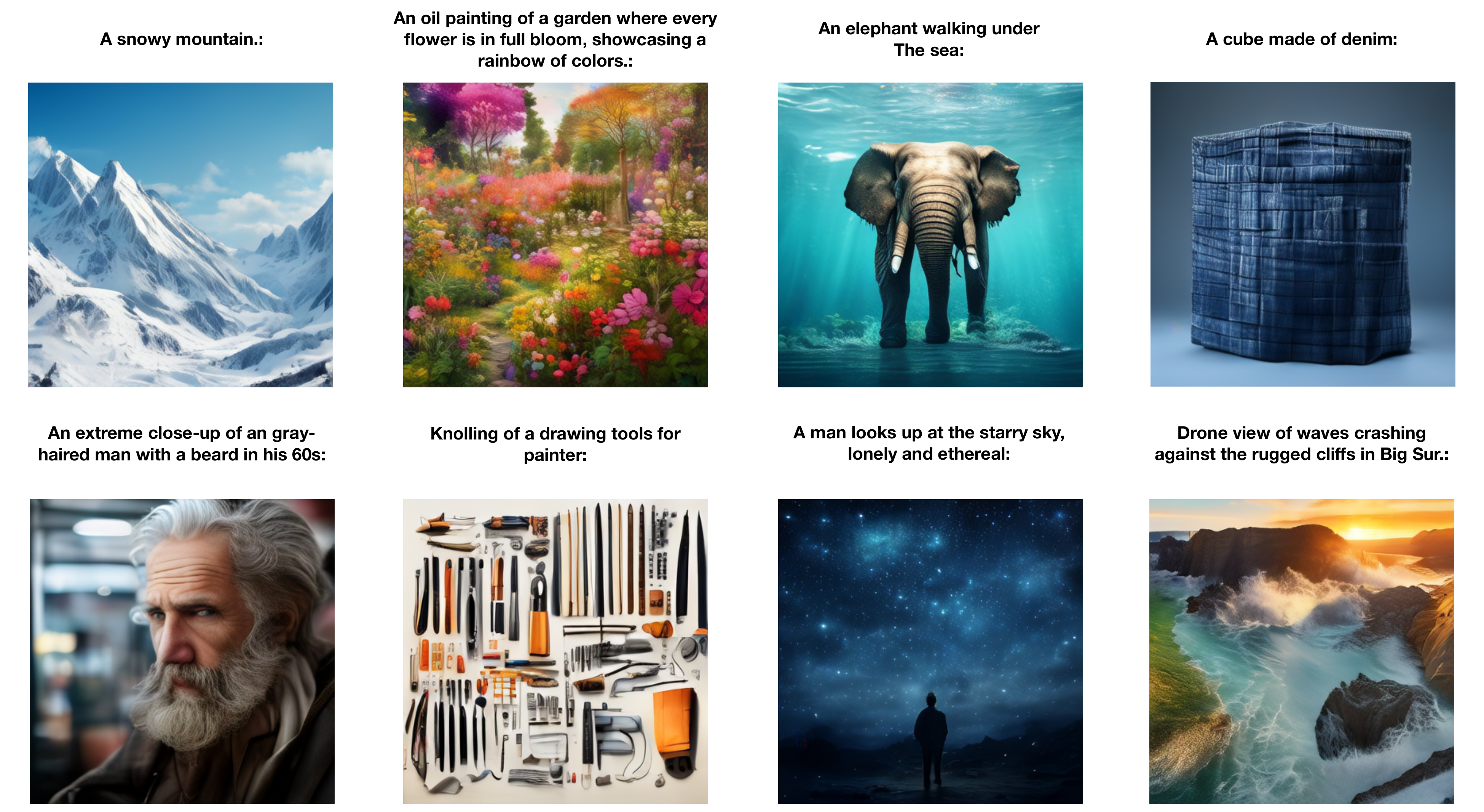}
    \caption{Image generation results. VILA-U can generate high-quality images given text input.}
    \label{fig:generation_more}
    \vspace{-15pt}
\end{figure}

\begin{figure}[htbp]
    \centering
    \includegraphics[width=\textwidth]{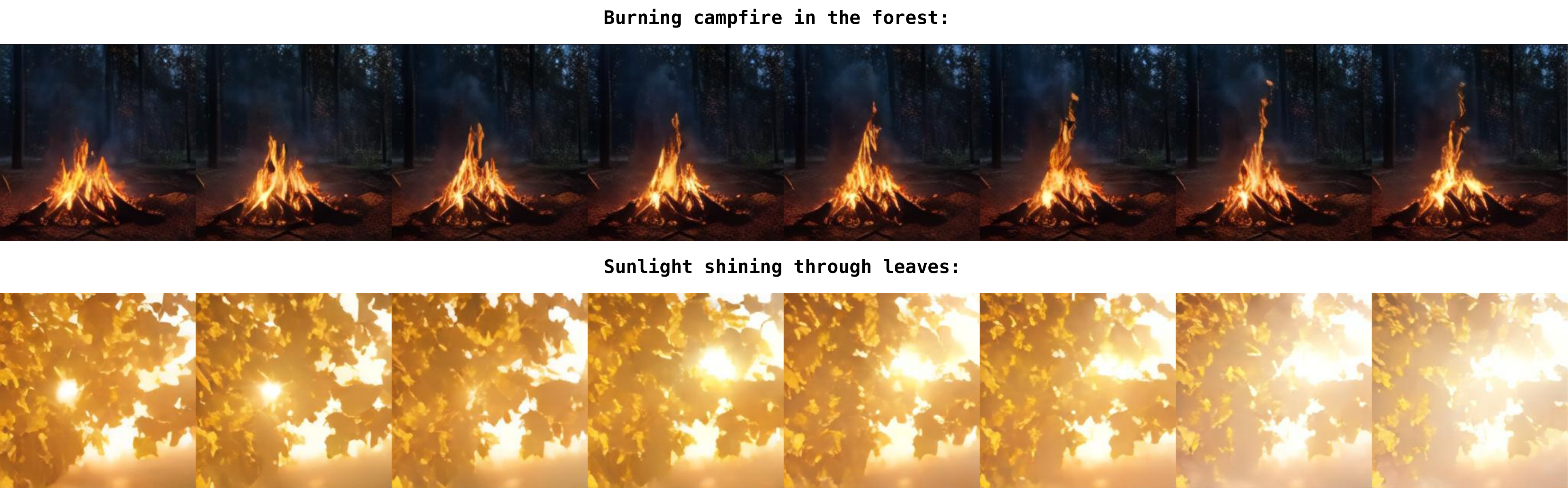}
    \caption{Video generation results. VILA-U can generate high-quality videos given text input.}
    \label{fig:video_more}
\end{figure}

We provide more image generation and video generation examples in Figure~\ref{fig:generation_more} and Figure~\ref{fig:video_more}. VILA-U can generate high-quality images and videos given text input.



\section{Failed Training Recipes.}
\label{appendix:C}

We experiment with numerous training recipes and find none to be as effective as our final approach. We list four alternative recipes and discuss their shortcomings compared to our final recipe: 1) Load pre-trained CLIP weights into the text encoder only; 2) Load pre-trained RQ-VAE weights for the vision encoder and decoder while training other parts from scratch; 3) Freeze the vision encoder; 4) Make the text encoder trainable.

Recipes 1) and 2) fail due to the lack of pre-trained CLIP weights for the vision encoder. Training a CLIP model from scratch typically requires numerous GPU days with a large global batch size (e.g., 32k). However, VQ-based reconstruction training necessitates a relatively small global batch size (e.g., 512) for steady improvement. With such a small batch size, training a text-aligned vision tower from scratch would be prohibitively time-consuming and resource-intensive.

Recipe 3) fails because freezing the vision encoder prevents it from learning the low-level features essential for reconstruction. In this case, the burden of reconstruction falls entirely on the vision decoder, but it is impossible to reconstruct images well using only semantic features.

Recipe 4) fails because the quantized features are chaotic during the initial training steps, and the contrastive loss disrupts the text encoder weights, slowing down the entire training process.

In contrast, our final training recipe leverages pre-trained CLIP weights for the vision encoder, enabling it to maintain learned semantic features rather than grasping them from scratch. This allows us to train with a small batch size while keeping the vision encoder trainable, facilitating the learning of low-level features for reconstruction during training.

\end{document}